\documentclass[10pt,twocolumn,letterpaper]{article}

\usepackage{cvpr}
\usepackage{times}
\usepackage{epsfig}
\usepackage{graphicx}
\usepackage{amsmath}
\usepackage{amssymb}
\usepackage{subfigure}

\graphicspath{{./graphics/}}
\usepackage{tabularx}
\usepackage{booktabs}
\usepackage{caption}
\usepackage{tikz}
\usepackage{pgfplots}
\usepackage{pgfplotstable}
\usepackage{standalone}
\usepackage{bm}
\usepackage[normalem]{ulem}
\newlength{\itemwidth}

\newcolumntype{Y}{>{\centering\arraybackslash}X}

\def\smallcheck{\tikz\fill[scale=0.25](0,.35) -- (.25,0) -- (1,.7) -- (.25,.15) -- cycle;}

\usetikzlibrary{calc}
\usetikzlibrary{tikzmark}

\pgfplotsset{compat=newest}
\definecolor{999933}{RGB}{153, 153, 51}
\definecolor{BEBADA}{RGB}{190,186,218}
\definecolor{FB8072}{RGB}{251,128,114}
\definecolor{80B1D3}{RGB}{128,177,211}
\definecolor{FDB463}{RGB}{253,180,98}
\definecolor{8DD3C7}{RGB}{141,211,199}
\pgfplotscreateplotcyclelist{custompalette}{
    {999933!50!black,fill=999933},
    {BEBADA!50!black,fill=BEBADA},
    {FB8072!50!black,fill=FB8072},
    {80B1D3!50!black,fill=80B1D3},
    {FDB463!50!black,fill=FDB463},
    {8DD3C7!50!black,fill=8DD3C7}
}

\usepackage[pagebackref=true,breaklinks=true,letterpaper=true,colorlinks,bookmarks=false]{hyperref}

\cvprfinalcopy

\begin{document}

\title{Video Frame Interpolation via Adaptive Convolution}

\author{Simon Niklaus\thanks{The first two authors contributed equally to this paper.}\\
Portland State University\\
{\tt\small sniklaus@pdx.edu}
\and
Long Mai\footnotemark[1]\\
Portland State University\\
{\tt\small mtlong@cs.pdx.edu}
\and
Feng Liu\\
Portland State University\\
{\tt\small fliu@cs.pdx.edu}
}

\maketitle

\begin{abstract}

   Video frame interpolation typically involves two  steps: motion estimation and pixel synthesis. Such a two-step approach heavily depends on the quality of motion estimation. This paper presents a robust video frame interpolation method that combines these two steps into a single process. Specifically, our method considers pixel synthesis for the interpolated frame as local convolution over two input frames. The convolution kernel captures both the local motion between the input frames and the coefficients for pixel synthesis. Our method employs a deep fully convolutional neural network to estimate a spatially-adaptive convolution kernel for each pixel. This deep neural network can be directly trained end to end using widely available video data without any difficult-to-obtain ground-truth data like optical flow. Our experiments show that the formulation of video interpolation as a single convolution process allows our method to gracefully handle challenges like occlusion, blur, and abrupt brightness change and enables high-quality video frame interpolation.

\end{abstract}

\vspace{-0.2in}
\section{Introduction}
\label{sec:intro}
Frame interpolation is a classic computer vision problem and is important for applications like novel view interpolation and frame rate conversion~\cite{Meyer_CVPR_2015}. Traditional frame interpolation methods have two steps: motion estimation, usually optical flow, and pixel synthesis~\cite{Baker_OTHER_2011}. Optical flow is often difficult to estimate in the regions suffering from occlusion, blur, and abrupt brightness change. Flow-based pixel synthesis cannot reliably handle the occlusion problem. Failure of any of these two steps will lead to noticeable artifacts in interpolated video frames.

\begin{figure}\centering
    \hspace*{-0.2cm}\includegraphics[]{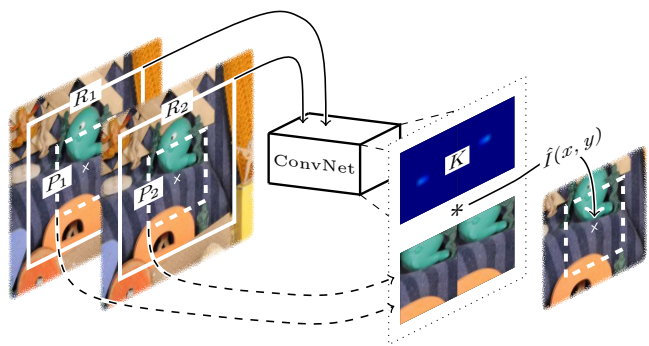}
	\caption{Pixel interpolation by convolution. For each output pixel $(x, y)$, our method estimates a convolution kernel $K$ and uses it to convolve with patches $P_1$ and $P_2$ centered at $(x, y)$ in the input frames to produce its color $\hat{I}(x,y)$.}\vspace{-0.2in}
	\label{fig:pipeline}
\end{figure}

This paper presents a robust video frame interpolation method that achieves frame interpolation using a deep convolutional neural network without explicitly dividing it into separate steps. Our method considers pixel interpolation as convolution over corresponding image patches in the two input video frames, and estimates the spatially-adaptive convolutional kernel using a deep fully convolutional neural network. Specifically, for a pixel $(x, y)$ in the interpolated frame, this deep neural network takes two receptive field patches $R_1$ and $R_2$ centered at that pixel as input and estimates a convolution kernel $K$. This convolution kernel is used to convolve with the input patches $P_1$ and $P_2$ to synthesize the output pixel, as illustrated in Figure~\ref{fig:pipeline}. \let\thefootnote\relax\footnote{\url{http://graphics.cs.pdx.edu/project/adaconv}}

An important aspect of our method is the formulation of pixel interpolation as convolution over pixel patches instead of relying on optical flow. This convolution formulation unifies motion estimation and pixel synthesis into a single procedure. It enables us to design a deep fully convolutional neural network for video frame interpolation  without dividing interpolation into separate steps. This formulation is also more flexible than those based on optical flow and can better handle challenging scenarios for frame interpolation. Furthermore, our neural network is able to estimate edge-aware convolution kernels that lead to sharp results.

The main contribution of this paper is a robust video frame interpolation method that employs a fully deep convolutional neural network to produce high-quality video interpolation results. This method has a few advantages. First, since it models video interpolation as a single process, it is able to make proper trade-offs among competing constraints and thus can provide a robust interpolation approach. Second, this frame interpolation deep convolutional neural network can be directly trained end to end using widely available video data, without any difficult-to-obtain ground truth data like optical flow. Third, as demonstrated in our experiments, our method can generate high-quality frame interpolation results for challenging videos such as those with occlusion, blurring artifacts, and abrupt brightness change.

\section{Related Work}
\label{sec:related}
Frame interpolation for video is one of the basic computer vision and video processing technologies. It is a special case of image-based rendering where middle frames are interpolated from temporally neighboring frames. Good surveys on image-based rendering are available~\cite{Kang_OTHER_2006, Szeliski_BOOK_2010, Zhang_OTHER_2004}. This section focuses on research that is specific to video frame interpolation and our work.

Most existing frame interpolation methods estimate dense motion between two consecutive input frames using stereo matching or optical flow algorithms and then interpolate one or more middle frames according to the estimated dense correspondences~\cite{Baker_OTHER_2011, Werlberger_OTHER_2011, Yu_OTHER_2013}. Different from these methods, Mahajan~\etal developed a moving gradient method that estimates paths in input images, copies proper gradients to each pixel in the frame to be interpolated and then synthesizes the interpolated frame via Poisson reconstruction~\cite{Mahajan_TOG_2009}. The performance of all the above methods depends on the quality of dense correspondence estimation and special care needs to be taken to handle issues like occlusion during the late image synthesis step. 

As an alternative to explicit motion estimation-based methods, phase-based methods have recently been shown promising for video processing. These methods encode motion in the phase difference between input frames and manipulate the phase information for applications like motion magnification~\cite{Wadhwa_TOG_2013} and view expansion~\cite{Didyk_TOG_2013}. Meyer~\etal further extended these approaches to accommodate large motion by propagating phase information across oriented multi-scale pyramid levels using a bounded shift correction strategy~\cite{Meyer_CVPR_2015}. This phase-based interpolation method can generate impressive video interpolation results and handle challenging scenarios gracefully; however, further improvement is still required to better preserve high-frequency detail in the video with large inter-frame changes.

Our work is inspired by the success of deep learning in solving not only difficult visual understanding problems~\cite{Girshick_CVPR_2014, He_CVPR_2016, Karayev_BMVC_2014, Krizhevsky_NIPS_2012, Sermanet_ICLR_2014, Simonyan_CORR_2014, Yisun_CVPR_2015, Taigman_CVPR_2014, Wu_CVPR_2015, Yosinski_NIPS_2014, Zhou_NIPS_2014} but also other computer vision problems like optical flow estimation~\cite{ Dosovitskiy_ICCV_2015, Gadot_CVPR_2015, Guney_ACCV_2016, Teney_CORR_2016, Tran_CVPR_2015, Weinzaepfel_ICCV_2013}, style transfer~\cite{Dumoulin_CORR_2016, Gatys_CVPR_2016, Johnson_ECCV_2016, Li_CVPR_2016,  Ulyanov_ICML_2016}, and image enhancement~\cite{Burger_CVPR_2012, Dong_ICCV_2015, Dong_PAMI_2016, Jiansun_CVPR_2015, Svoboda_CORR_2016, Xie_NIPS_2012, Xu_NIPS_2014, Zhang_ECCV_2016, Zhu_ECCV_2016}. Our method is particularly relevant to the recent deep learning algorithms for view synthesis~\cite{Dosovitskiy_CVPR_2015, Flynn_CVPR_2016, Kalantari_TOG_2016, Kulkarni_NIPS_2015, Tatarchenko_ECCV_2016, Yang_NIPS_2015, Zhou_ECCV_2016}. Dosovitiskiy~\etal~\cite{Dosovitskiy_CVPR_2015}, Kulkarni~\etal~\cite{Kulkarni_NIPS_2015}, Yang~\etal~\cite{Yang_NIPS_2015}, and Tatarchenko~\etal~\cite{Tatarchenko_ECCV_2016} developed deep learning algorithms that can render unseen views from input images. These algorithms work on objects, such as chairs and faces, and are not designed for frame interpolation for videos of general scenes.

Recently, Flynn~\etal developed a deep convolutional neural network method for synthesizing novel natural images from posed real-world input images. Their method projects input images onto multiple depth planes and combines colors at these depth planes to create a novel view~\cite{Flynn_CVPR_2016}. Kalantari~\etal provided a deep learning-based view synthesis algorithm for view expansion for light field imaging. They break novel synthesis into two components: disparity and color estimation, and accordingly use two sequential convolutional neural networks to model these two components. These two neural networks are trained simultaneously~\cite{Kalantari_TOG_2016}. Long~\etal interpolate frames as an intermediate step for image matching~\cite{Long_ECCV_2016}. However, their interpolated frames tend to be blurry. Zhou~\etal observed that the visual appearance of different views of the same instance is highly correlated, and designed a deep learning algorithm to predict appearance flows that are used to select proper pixels in the input views to synthesize a novel view~\cite{Zhou_ECCV_2016}. Given multiple input views, their method can interpolate a novel view by warping individual input views using the corresponding appearance flows and then properly combining them together. Like these methods, our deep learning algorithm can also be trained end to end using videos directly. Compared to these methods, our method is dedicated to video frame interpolation. More importantly, our method estimates convolution kernels that capture both the motion and interpolation coefficients, and uses these kernels to directly convolve with input images to synthesize a middle video frame. Our method does not need to project input images onto multiple depth planes or explicitly estimate disparities or appearance flows to warp input images and then combine them together. Our experiments show that our formulation of frame interpolation as a single convolution step allows our method to robustly handle challenging cases. Finally, the idea of using convolution for image synthesis has also been explored in the very recent work for frame extrapolation~\cite{Finn_NIPS_2016, Jia_NIPS_2016, Xue_NIPS_2016}.

\section{Video Frame Interpolation}
\label{sec:method}
\begin{figure}\centering
    \subfigure[Interpolation by motion estimation and color interpolation]{\includegraphics[]{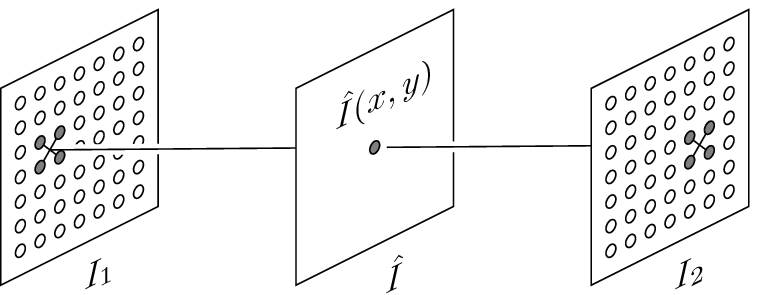}}
    \subfigure[Interpolation by convolution]{\includegraphics[]{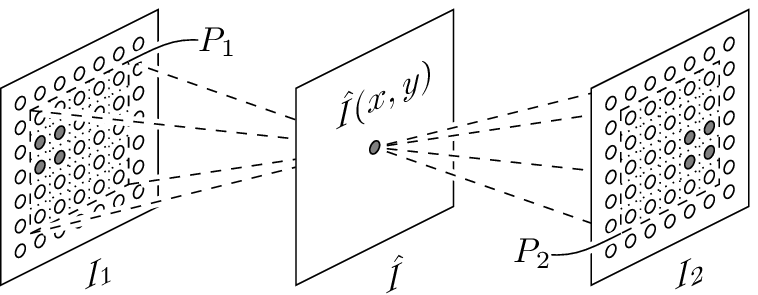}}\vspace{-0.1in}
	\caption{Interpolation by convolution. (a): a two-step approach first estimates motion between two frames and then interpolates the pixel color based on the motion. (b): our method directly estimates a convolution kernel and uses it to convolve the two frames to interpolate the pixel color.}\vspace{-0.25in}
	\label{fig:interpolation}
\end{figure}

Given two video frames $I_1$ and $I_2$, our method aims to interpolate a frame $\hat{I}$ temporally in the middle of the two input frames. Traditional interpolation methods estimate the color of a pixel $\hat{I}(x, y)$ in the interpolated frame in two steps: dense motion estimation, typically through optical flow, and pixel interpolation. For instance, we can find for pixel $(x, y)$ its corresponding pixels $(x_1, y_1)$ in $I_1$ and $(x_2, y_2)$ in $I_2$ and then interpolate the color from these corresponding pixels. Often this step also involves re-sampling images $I_1$ and $I_2$ to obtain the corresponding values $I_1(x_1, y_1)$ and $I_2(x_2, y_2)$ to produce a high-quality interpolation result, especially when $(x_1, y_1)$ and $(x_2, y_2)$ are not integer locations, as illustrated in Figure~\ref{fig:interpolation} (a). This two-step approach can be compromised when optical flow is not reliable due to occlusion, motion blur, and lack of texture. Also, rounding the coordinates to find the color for $I_1(x_1, y_1)$ and $I_2(x_2, y_2)$ is prone to aliasing while re-sampling with a fixed kernel sometimes cannot preserve sharp edges well. Advanced re-sampling methods exist and can be used for edge-preserving re-sampling, which, however, requires high-quality optical flow estimation.

Our solution is to combine motion estimation and pixel synthesis into a single step and formulate pixel interpolation as a local convolution over patches in the input images $I_1$ and $I_2$. As shown in Figure~\ref{fig:interpolation} (b), the color of pixel $(x, y)$ in the target image to be interpolated can be obtained by convolving a proper kernel $K$ over input patches $P_1(x, y)$ and $P_2(x, y)$ that are also centered at $(x, y)$ in the respective input images. The convolutional kernel $K$ captures both motion and re-sampling coefficients for pixel synthesis. This formulation of pixel interpolation as convolution has a few advantages. First of all, the combination of motion estimation and pixel synthesis into a single step provides a more robust solution than the two-step procedure. Second, the convolution kernel provides flexibility to account for and address difficult cases like occlusion. For example, optical flow estimation in an occlusion region is a fundamentally difficult problem, which makes it difficult for a typical two-step approach to proceed. Extra steps based on heuristics, such as flow interpolation, must be taken. This paper provides a data-driven approach to directly estimate the convolution kernel that can produce visually plausible interpolation results for an occluded region. Third, if properly estimated, this convolution formulation can seamlessly integrate advanced re-sampling techniques like edge-aware filtering to provide sharp interpolation results.

Estimating proper convolution kernels is essential for our method. Encouraged by the success of using deep learning algorithms for optical flow estimation~\cite{Dosovitskiy_ICCV_2015, Gadot_CVPR_2015, Guney_ACCV_2016, Teney_CORR_2016, Tran_CVPR_2015, Weinzaepfel_ICCV_2013} and image synthesis~\cite{Flynn_CVPR_2016, Kalantari_TOG_2016, Zhou_ECCV_2016}, we develop a deep convolutional neural network method to estimate a proper convolutional kernel to synthesize each output pixel in the interpolated images. The convolutional kernels for individual pixels vary according to the local motion and image structure to provide high-quality interpolation results. Below we describe our deep neural network for kernel estimation and then discuss implementation details.

\begin{table}\centering\small
    \begin{tabularx}{\columnwidth}{X @{\hspace{0.2cm}} c @{\hspace{0.2cm}} c @{\hspace{0.2cm}} c @{\hspace{0.2cm}} c @{\hspace{0.2cm}} r @{\thinspace} c @{\thinspace} c @{\thinspace} c @{\thinspace} c}
        \toprule
            type & BN & ReLU & size & stride & \multicolumn{5}{c}{output}
        \\ \midrule
            input & - & - & - & - & $6$ & $\times$ & $79$ & $\times$ & $79$
        \\
            conv & \smallcheck & \smallcheck & $7 \times 7$ & $1 \times 1$ & $32$ & $\times$ & $73$ & $\times$ & $73$
        \\
            down-conv & - & \smallcheck & $2 \times 2$ & $2 \times 2$ & $32$ & $\times$ & $36$ & $\times$ & $36$
        \\
            conv & \smallcheck & \smallcheck & $5 \times 5$ & $1 \times 1$ & $64$ & $\times$ & $32$ & $\times$ & $32$
        \\
            down-conv & - & \smallcheck & $2 \times 2$ & $2 \times 2$ & $64$ & $\times$ & $16$ & $\times$ & $16$
        \\
            conv & \smallcheck & \smallcheck & $5 \times 5$ & $1 \times 1$ & $128$ & $\times$ & $12$ & $\times$ & $12$
        \\
            down-conv & - & \smallcheck & $2 \times 2$ & $2 \times 2$ & $128$ & $\times$ & $6$ & $\times$ & $6$
        \\
            conv & \smallcheck & \smallcheck & $3 \times 3$ & $1 \times 1$ & $256$ & $\times$ & $4$ & $\times$ & $4$
        \\
            conv & - & \smallcheck & $4 \times 4$ & $1 \times 1$ & $2048$ & $\times$ & $1$ & $\times$ & $1$
        \\
            conv & - & - & $1 \times 1$ & $1 \times 1$ & $3362$ & $\times$ & $1$ & $\times$ & $1$
        \\
            spatial softmax & - & - & - & - & $3362$ & $\times$ & $1$ & $\times$ & $1$
        \\
            output & - & - & - & - & $41 \! \times \! 82$ & $\times$ & $1$ & $\times$ & $1$
        \\ \bottomrule
    \end{tabularx}
    \normalsize\caption{The convolutional neural network architecture. It makes use of Batch Normalization (BN)~\cite{Sergey_ICML_2015} as well as Rectified Linear Units (ReLU). Note that the output only reshapes the result without altering its value.}\vspace{-0.2in}
    \label{tbl:arch}
\end{table}

\subsection{Convolution kernel estimation}
\label{sec:method:cnn}

We design a fully convolutional neural network to estimate the convolution kernels for individual output pixels. The architecture of our neural network is detailed in Table~\ref{tbl:arch}. Specifically, to estimate the convolutional kernel $K$ for the output pixel $(x, y)$, our neural network takes receptive field patches $R_1(x, y)$ and $R_2(x, y)$ as input. $R_1(x, y)$ and $R_2(x, y)$ are both centered at $(x, y)$ in the respective input images. The patches $P_1$ and $P_2$ that the output kernel will convolve in order to produce the color for the output pixel $(x, y)$ are co-centered at the same locations as these receptive fields, but with a smaller size, as illustrated in Figure~\ref{fig:pipeline}. We use a larger receptive field than the patch to better handle the aperture problem in motion estimation. In our implementation, the default receptive field size is $79 \times 79$ pixels. The convolution patch size is $41 \times 41$ and the kernel size is $41 \times 82$ as it is used to convolve with two patches. Our method applies the same convolution kernel to each of the three color channels.

As shown in Table~\ref{tbl:arch}, our convolutional neural network consists of several convolutional layers as well as down-convolutions as alternatives to max-pooling layers. We use Rectified Linear Units as activation functions and Batch Normalization~\cite{Sergey_ICML_2015} for regularization. We employ no further techniques for regularization since our neural network can be trained end to end using widely available video data, which provides a sufficiently large training dataset. We are also able to make use of data augmentation extensively, by horizontally and vertically flipping the training samples as well as reversing their order. Our neural network is fully convolutional. Therefore, it is not restricted to a fixed-size input and we are, as detailed in Section~\ref{sec:method:impl}, able to use a shift-and-stitch technique~\cite{Giusti_ICIP_2013, Long_CVPR_2015, Sermanet_ICLR_2014} to produce kernels for multiple pixels simultaneously to speedup our method.

A critical constraint is that the coefficients of the output convolution kernel should be non-negative and sum up to one. Therefore, we connect the final convolutional layer to a spatial softmax layer to output the convolution kernel, which implicitly meets this important constraint. 

\begin{figure}\centering
    \setlength{\tabcolsep}{0.0cm}
    \setlength{\itemwidth}{4.1cm}

    \begin{tabularx}{\textwidth}{c @{\hspace{0.05cm}} c}
            \includegraphics[width=\itemwidth,trim={8.5cm 4.5cm 5cm 7.5cm},clip]{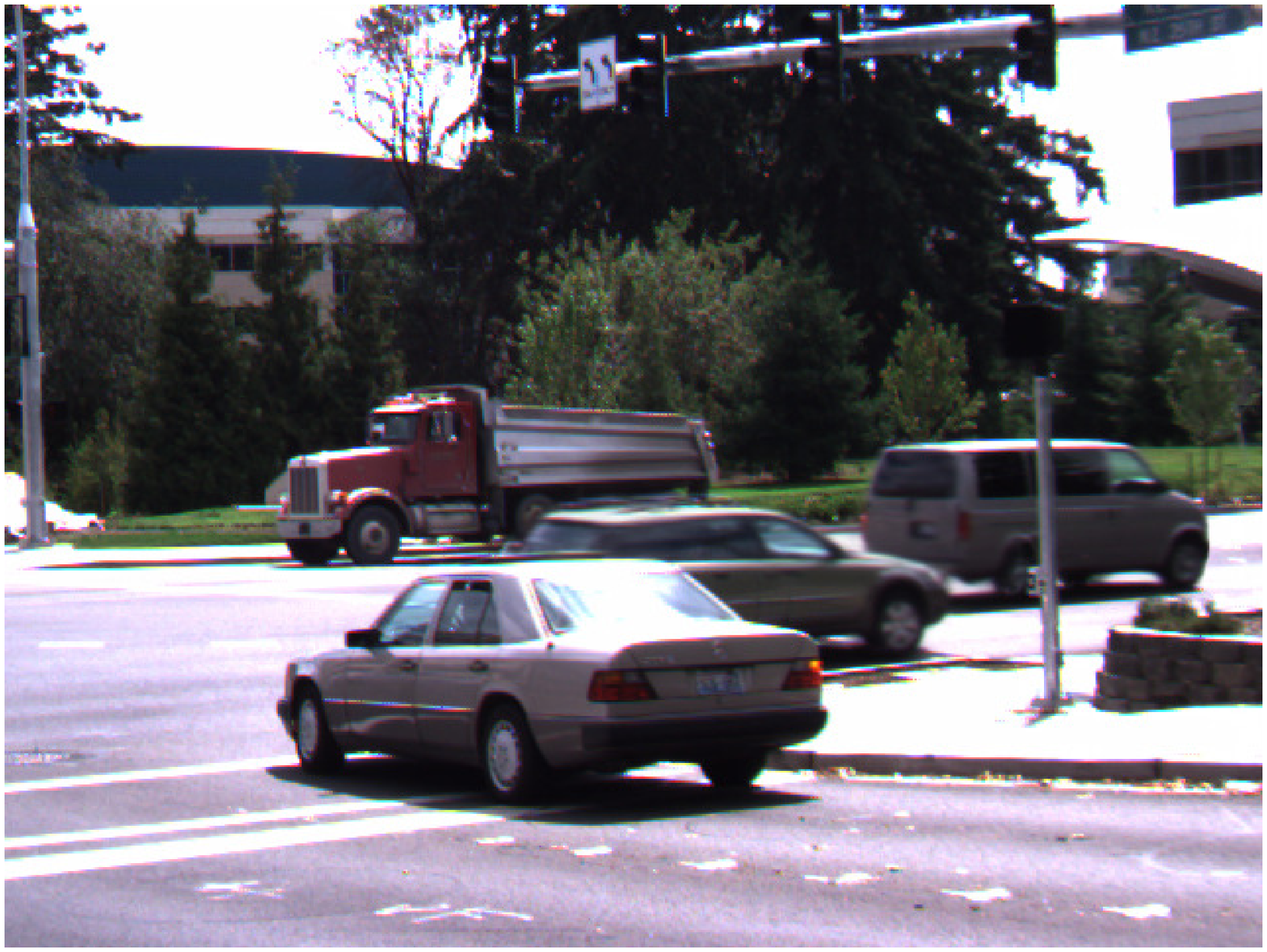}
        &
            \includegraphics[width=\itemwidth,trim={8.5cm 4.5cm 5cm 7.5cm},clip]{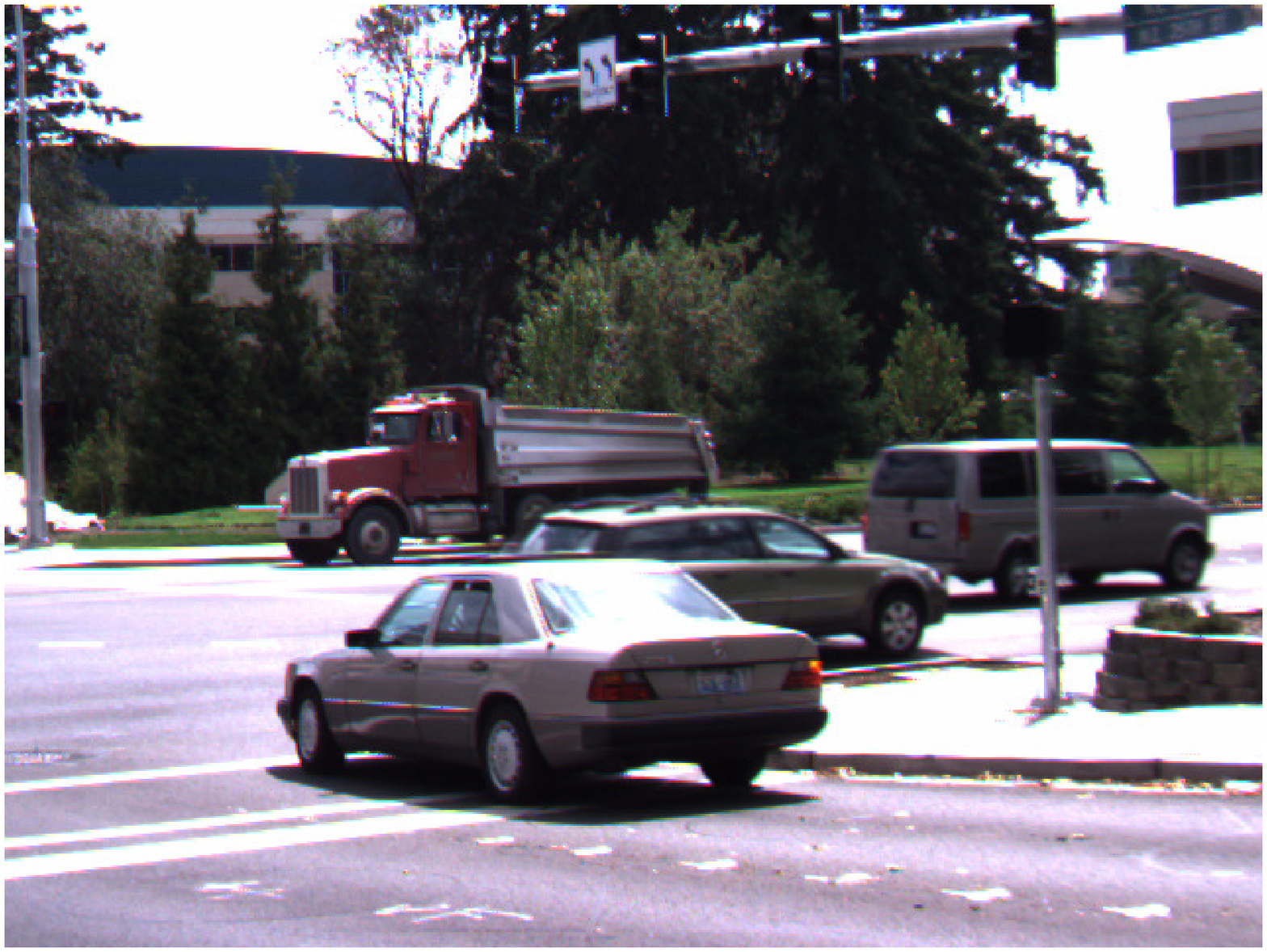}
        \vspace{-0.1cm} \\
             color loss
        &
             color loss + gradient loss
        \\
    \end{tabularx}\vspace{-0.1in}
    \caption{Effect of using an additional gradient loss.}\vspace{-0.2in}
    \label{fig:loss}
\end{figure}

\subsubsection{Loss function}

For clarity, we first define notations. The $i^{th}$ training example  consists of two input receptive field patches $R_{i,1}$ and $R_{i,2}$ centered at $(x_i, y_i)$, the corresponding input patches $P_{i,1}$ and $P_{i,2}$ that are smaller than the receptive field patches and also centered at the same location, the ground-truth color $\tilde{C}_i$ and the ground-truth gradient $\tilde{G}_i$ at $(x_i, y_i)$ in the interpolated frame. For simplicity, we omit the $(x_i, y_i)$ in our definition of the loss functions. 

One possible loss function of our deep convolutional neural network can be the difference between the interpolated pixel color and the ground-truth color as follows.
\begin{equation}
    E_c=\sum_i \|[P_{i,1} \; P_{i,2}]\ast K_i-\tilde{C}_i\|_1
\end{equation}
where subscript $i$ indicates the $i^{th}$ training example and $K_i$ is the convolution kernel output by our neural network. Our experiments show that this color loss alone, even using $\ell_1$ norm, can lead to blurry results, as shown in Figure~\ref{fig:loss}. This blurriness problem was also reported in some recent work~\cite{Long_ECCV_2016, Mathieu_ICLR_2016, Ranzato_CORR_2014}. Mathieu~\etal showed that this blurriness problem can be alleviated by incorporating image gradients in the loss function~\cite{Mathieu_ICLR_2016}. This is difficult within our pixel-wise interpolation approach, since the image gradient cannot be directly calculated from a single pixel. Since differentiation is also a convolution, assuming that kernels locally vary slowly, we solve this problem by using the associative property of convolution: we first compute the gradient of input patches and then perform convolution with the estimated kernel, which will result in the gradient of the interpolated image at the pixel of interest. As a pixel $(x, y)$ has eight immediate neighboring pixels, we compute eight versions of gradients using finite difference and incorporate all of them into our gradient loss function. 
\begin{equation}
    E_g=\sum_i \sum_{k=1}^8 \|[G^k_{i,1} \; G^k_{i,2}]\ast K_i-\tilde{G}^k_i\|_1
\end{equation}
where $k$ denotes one of the eight ways we compute the gradient. $G^k_{i,1}$ and $G^k_{i,2}$ are the gradients of the input patches $P_{i,1}$ and $P_{i,2}$, and $\tilde{G}^k_i$ is the ground-truth gradient. We combine the above color and gradient loss as our final loss $E_c + \lambda \cdot E_g$. We found that $\lambda = 1$ works well and used it. As shown in Figure~\ref{fig:loss}, this color plus gradient loss enables our method to produce sharper interpolation results.

\subsection{Training} 

We derived our training dataset from an online video collection, as detailed later on in this section. To train our neural network, we initialize its parameters using the Xavier initialization approach~\cite{Xavier_OTHER_2010} and then use AdaMax~\cite{Kingma_CORR_2014} with $\beta_1 = 0.9$, $\beta_2 = 0.999$, a learning rate of 0.001 and $128$ samples per mini-batch to minimize the loss function.

\subsubsection{Training dataset}

Our loss function is purely based on the ground truth video frame and does not need any other ground truth information like optical flow. Therefore, we can make use of videos that are widely available online to train our neural network. To make it easy to reproduce our results, we use publicly available videos from Flickr with a Creative Commons license. We downloaded $3,000$ videos using keywords, such as ``driving'', ``dancing'', ``surfing'', ``riding'', and ``skiing'', which yield a diverse selection. We scaled the downloaded videos to a fixed size of $1280 \times 720$ pixels. We removed interlaced videos that sometimes have a lower quality than the videos with the progressive-scan format.

To generate the training samples, we group all the frames in each of the remaining videos into triple-frame groups, each containing three consecutive frames in a video. We then randomly pick a pixel in each triple-frame group and extract a triple-patch group centered at that pixel from the video frames. To facilitate data augmentation, the patches are selected to be larger than the receptive-field patches required by the neural network. The patch size in our training dataset is $150 \times 150$ pixels. To avoid including a large number of samples with no or little motion, we estimate the optical flow between patches from the first and last frame in the triple-frame group~\cite{Tao_OTHER_2012} and compute the mean flow magnitude. We then sample $500,000$ triple-patch groups without replacement according to the flow magnitude: a patch group with larger motion is more likely to be chosen than the one with smaller motion. In this way, our training set includes samples with a wide range of motion while avoiding being dominated by patches with little motion. Since some videos consist of many shots, we compute the color histogram between patches to detect shot boundaries and remove the groups across the shot boundaries. Furthermore, samples with little texture are also not very useful to train our neural network. We therefore compute the entropy of patches in each sample and finally select the $250,000$ triple-patch groups with the largest entropy to form the training dataset. In this training dataset, about $10$ percent of the pixels have an estimated flow magnitude of at least $20$ pixels. The average magnitude of the largest five percent is approximately $25$ pixels and the largest magnitude is $38$ pixels.

We perform data augmentation on the fly during training. The receptive-field size required for the neural network is $79\times 79$, which is smaller than the patch size in the training samples. Therefore, during the training, we randomly crop the receptive field patch from each training sample. We furthermore randomly flip the samples horizontally as well as vertically and randomly swap their temporal order. This forces the optical flow within the samples to be distributed symmetrically so that the neural network is not biased towards a certain direction.

\subsection{Implementation details}
\label{sec:method:impl}

We used Torch~\cite{Collobert_OTHER_2011} to implemented our neural network. Below we describe some important details.

\vspace{-0.1in}
\subsubsection{Shift-and-stitch implementation}

A straightforward way to apply our neural network to frame interpolation is to estimate the convolution kernel and synthesize the interpolated pixel one by one. This pixel-wise application of our neural network will unnecessarily perform redundant computations when passing two neighboring pairs of patches through the neural network to estimate the convolution kernels for two corresponding pixels. Our implementation employs the shift-and-stitch approach to address this problem to speedup our system~\cite{Giusti_ICIP_2013, Long_CVPR_2015, Sermanet_ICLR_2014}. 

Specifically, as our neural network is fully convolutional and does not require a fixed-size input, it can compute kernels for more than one output pixels at once by supplying a larger input than what is required to produce one kernel. This can mitigate the issue of redundant computations. The output pixels that are obtained in this way are however not adjacent and are instead sparsely distributed. We employ the shift-and-stitch~\cite{Giusti_ICIP_2013, Long_CVPR_2015, Sermanet_ICLR_2014} approach in which slightly shifted versions of the same input are used. This approach returns sparse results that can be combined to form the dense representation of the interpolated frame. 

Considering a frame with size $1280 \times 720$, a pixel-wise implementation of our neural network would require 921,600 forward passes through our neural network. The shift-and-stitch implementation of our neural network only requires 64 forward passes for the 64 differently shifted versions of the input to cope with the downscaling by the three down-convolutions. Compared to the pixel-wise implementation that takes $104$ seconds per frame on an Nvidia Titan X, the shift-and-stitch implementation only takes $9$ seconds.

\vspace{-0.1in}
\subsubsection{Boundary handling}

Due to the receptive field of the network as well as the size of the convolution kernel, we need to pad the input frames to synthesize boundary pixels for the interpolated frame. In our implementation, we adopt zero-padding. Our experiments show that this approach usually works well and does not introduce noticeable artifacts.

\vspace{-0.1in}
\subsubsection{Hyper-parameter selection}

The convolution kernel size and the receptive field size are two important hyper-parameters of our deep neural network. In theory, the convolution kernel, as shown in Figure~\ref{fig:interpolation}, must be larger than the pixel motion between two frames in order to capture the motion (implicitly) to produce a good interpolation result. To make our neural network robust against large motion, we tend to choose a large kernel. On the other hand, a large kernel involves a large number of values to be estimated, which increases the complexity of our neural network. We choose to select a convolution kernel that is large enough to capture the largest motion in the training dataset, which is 38 pixels. Particularly, the convolution kernel size in our system is $41 \times 82$ that will be applied to two $41 \times 41$ patches as illustrated in Figure~\ref{fig:pipeline}. We make this kernel a few pixels larger than 38 pixels to provide pixel support for re-sampling, which our method does not explicitly perform, but is captured in the kernel.

As discussed earlier, the receptive field is larger than the convolution kernel to handle the aperture problem well. However, a larger receptive field requires more computation and is less sensitive to the motion. We choose the receptive field using a validation dataset and find that $79 \times 79$ achieves a good balance.

\section{Experiments}
\label{sec:exp}
\begin{figure*}\centering
    \setlength{\tabcolsep}{0.0cm}
    \setlength{\itemwidth}{2.45cm}

    \begin{tabularx}{\textwidth}{c @{\hspace{0.05cm}} c @{\hspace{0.05cm}} c @{\hspace{0.05cm}} c @{\hspace{0.05cm}} c @{\hspace{0.05cm}} c @{\hspace{0.05cm}} c}
            \includegraphics[width=\itemwidth]{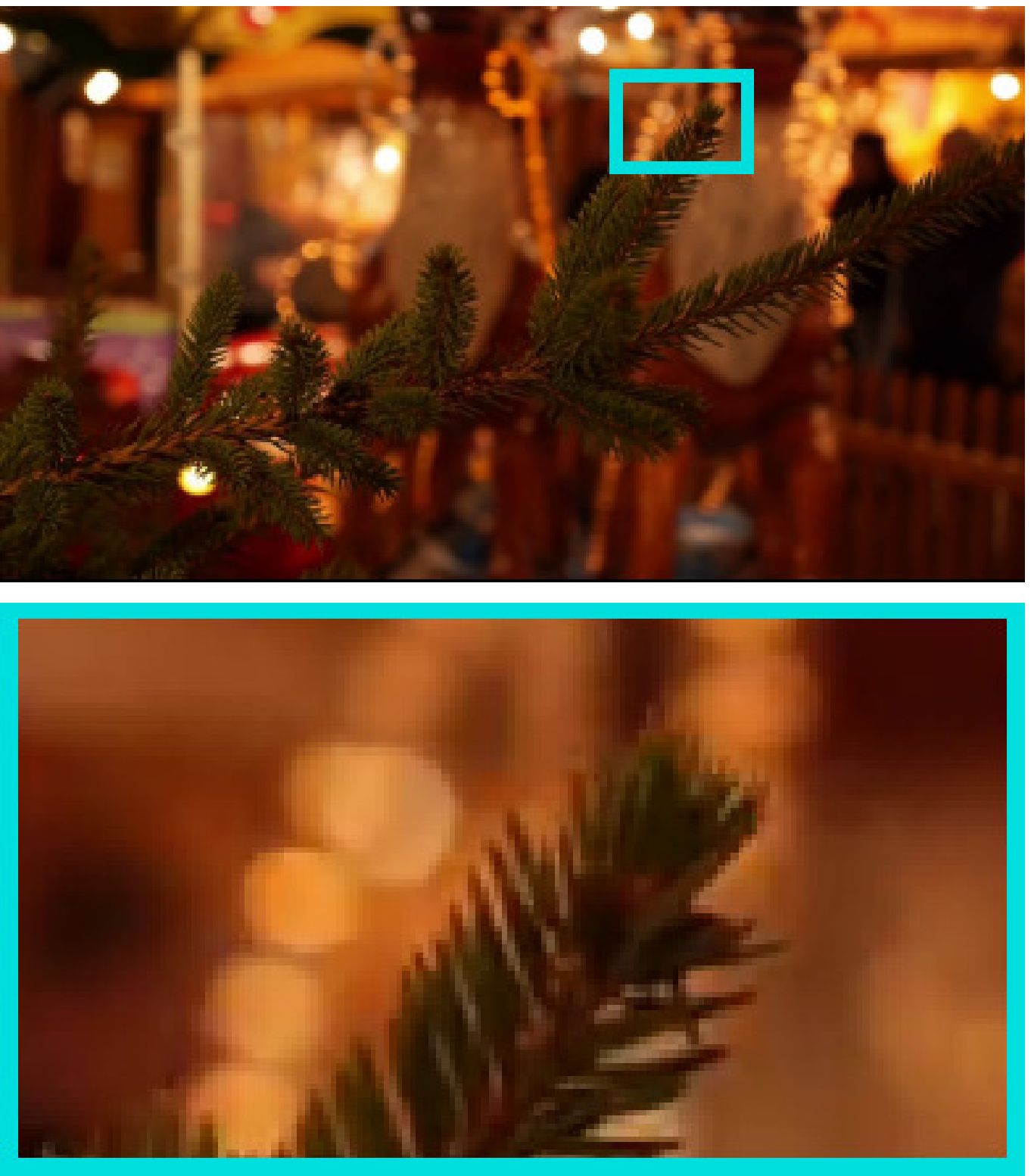}
        &
            \includegraphics[width=\itemwidth]{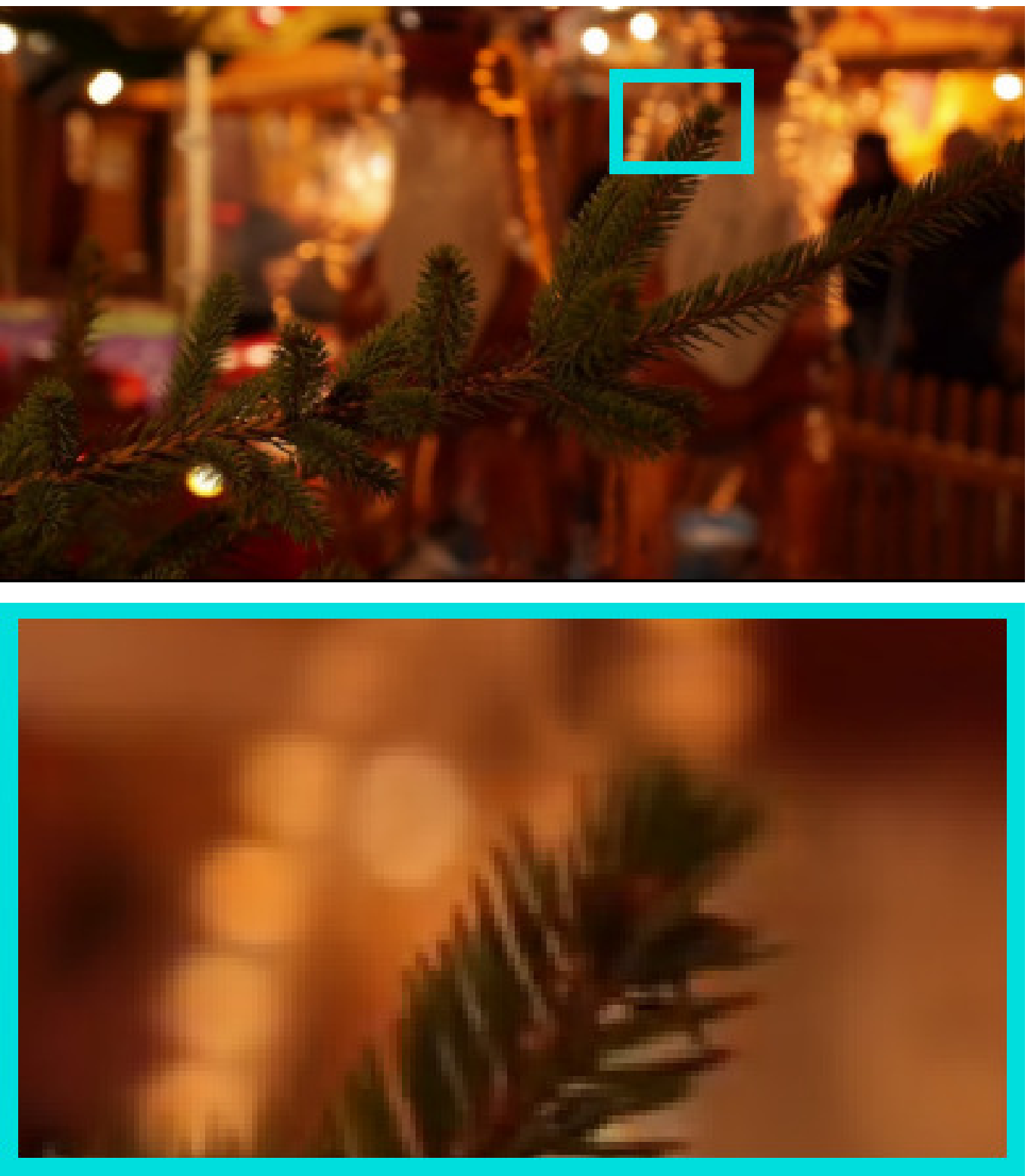}
        &
            \includegraphics[width=\itemwidth]{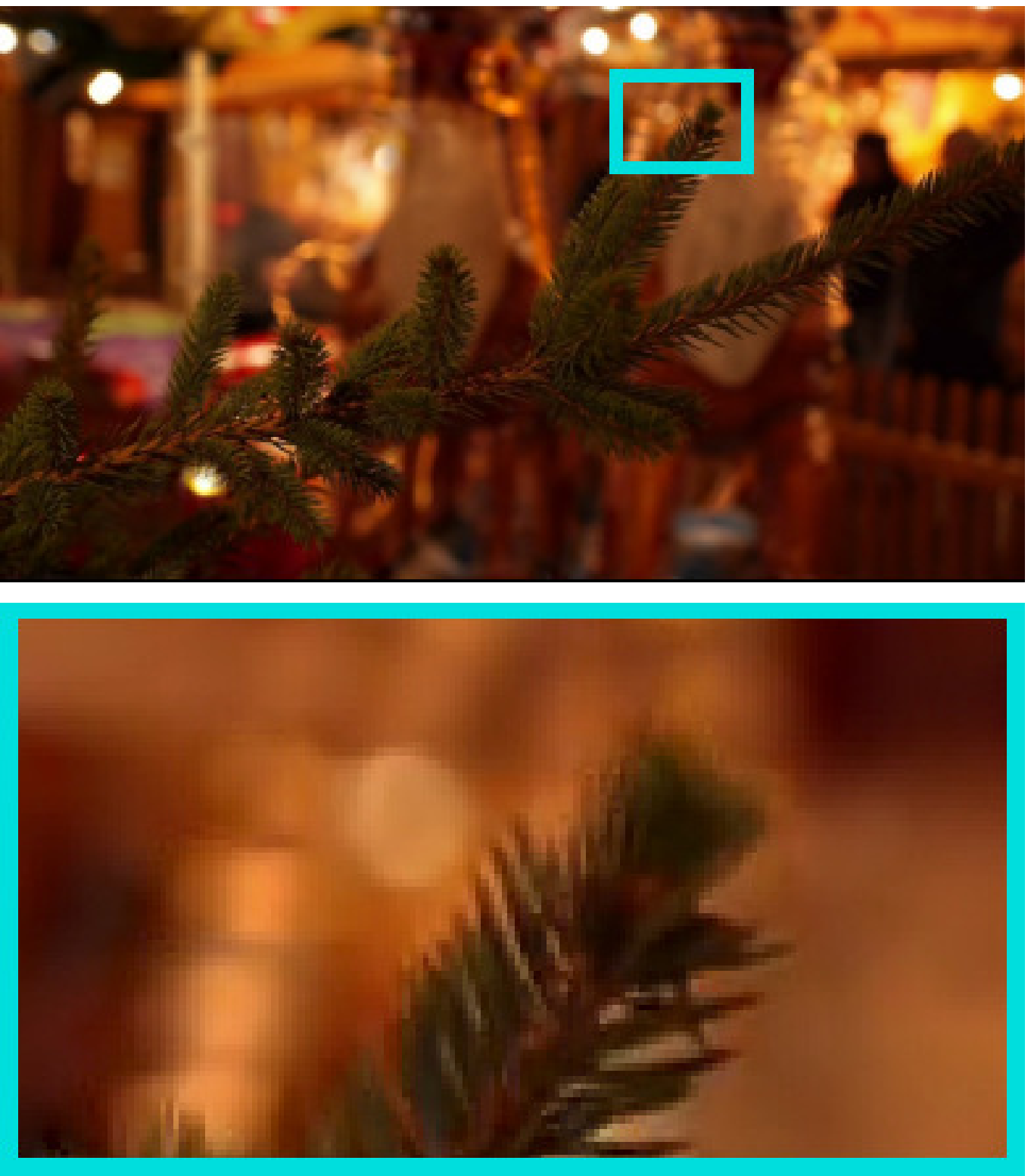}
        &
            \includegraphics[width=\itemwidth]{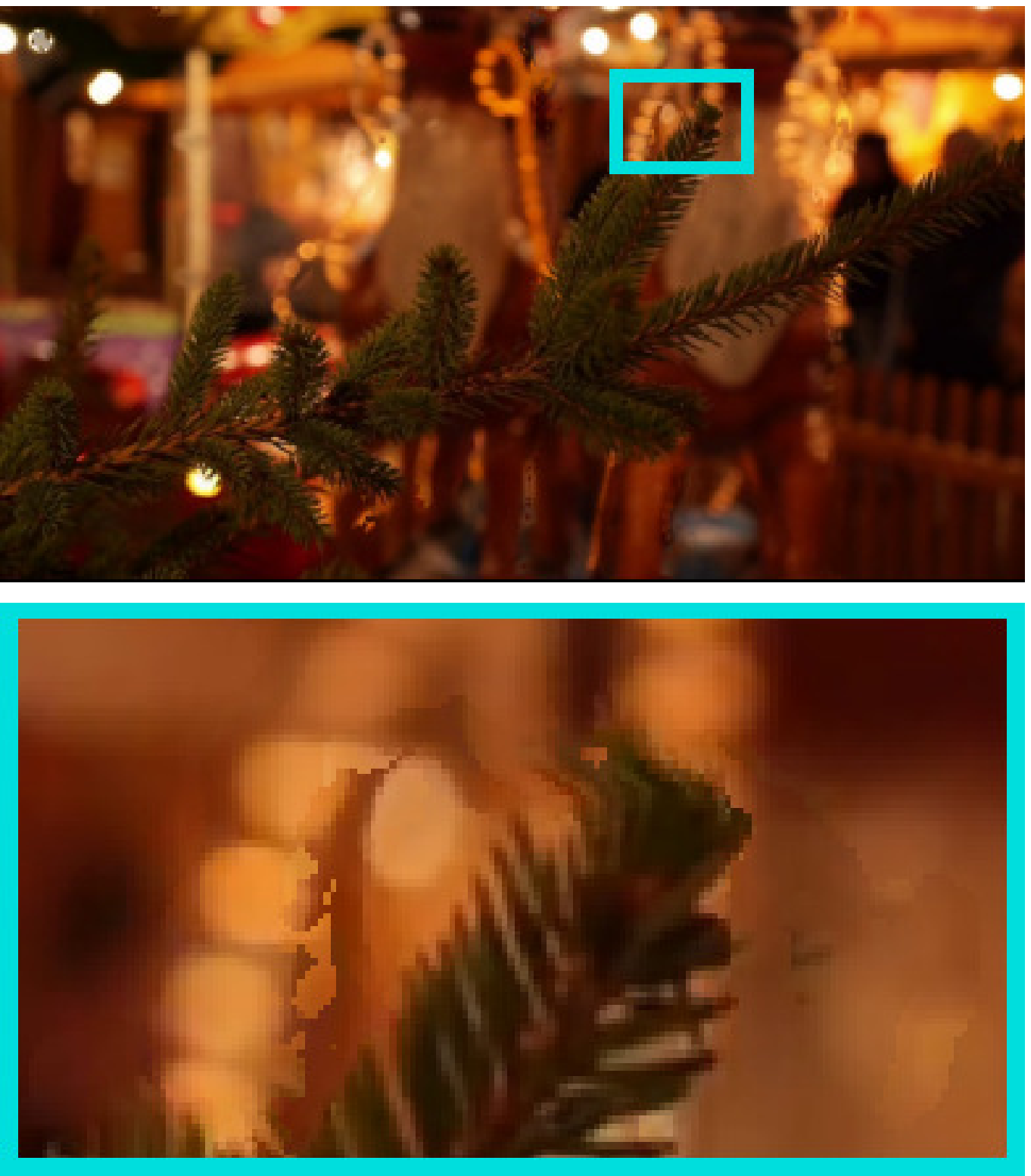}
        &
            \includegraphics[width=\itemwidth]{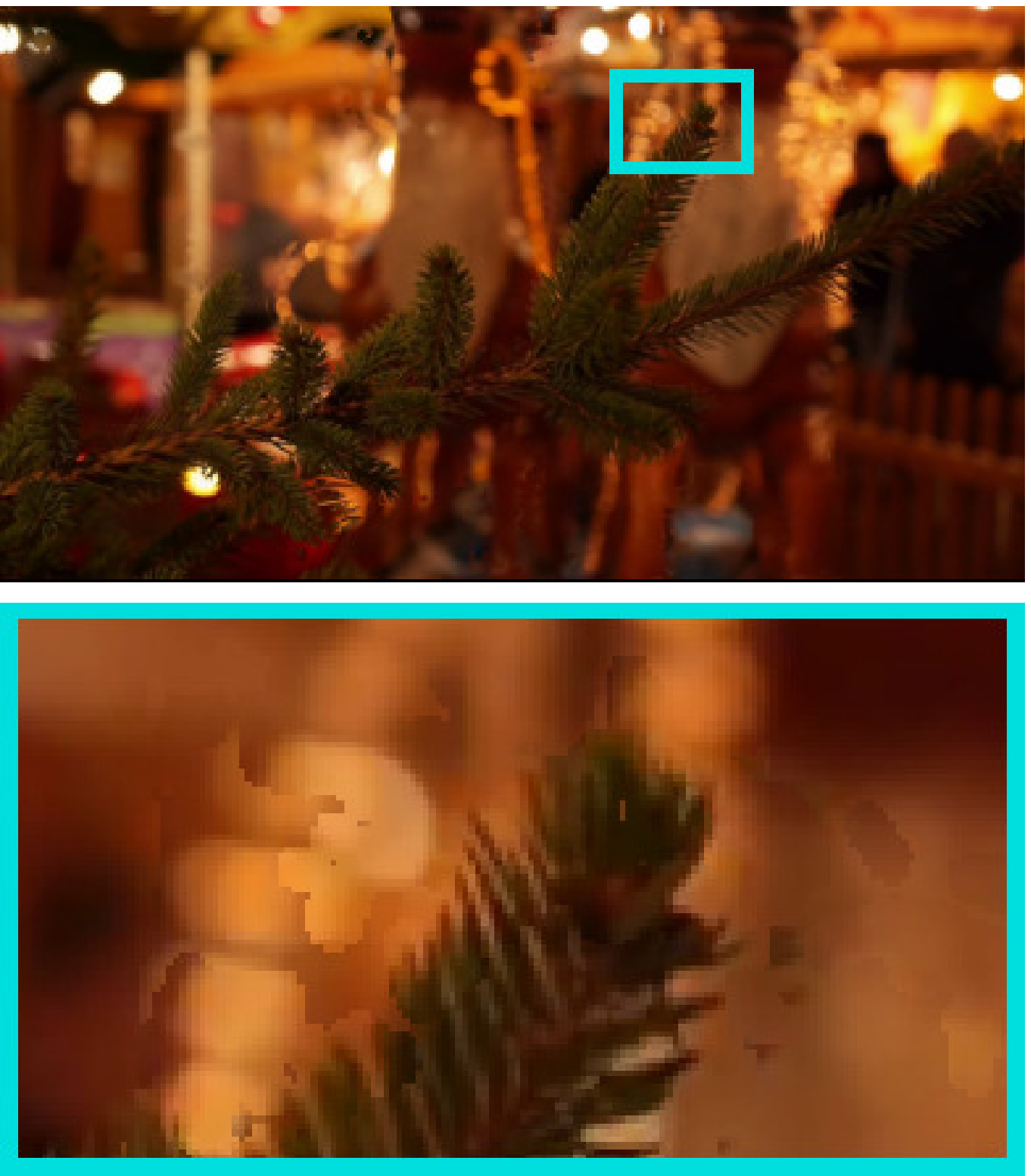}
        &
            \includegraphics[width=\itemwidth]{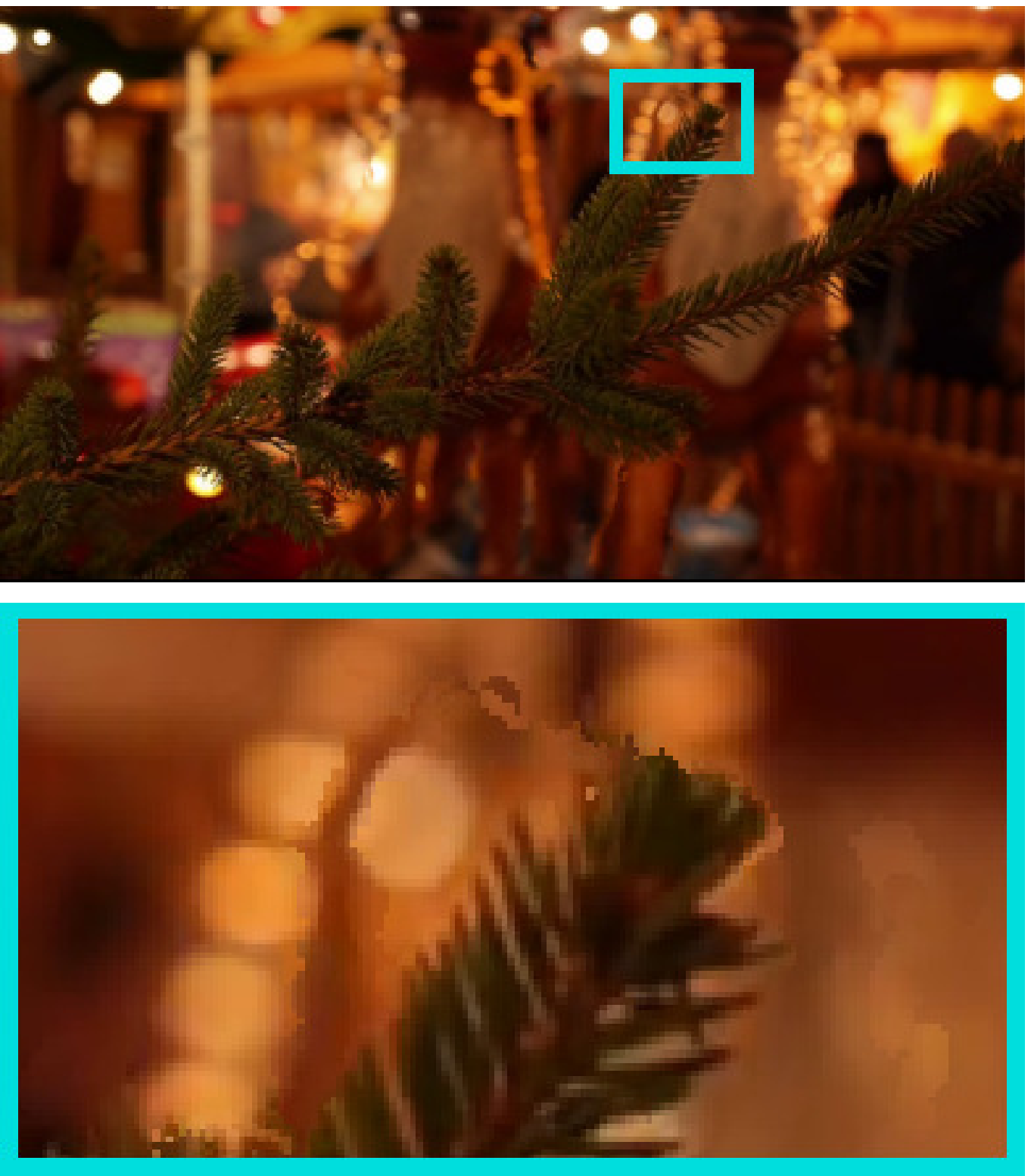}
        &
            \includegraphics[width=\itemwidth]{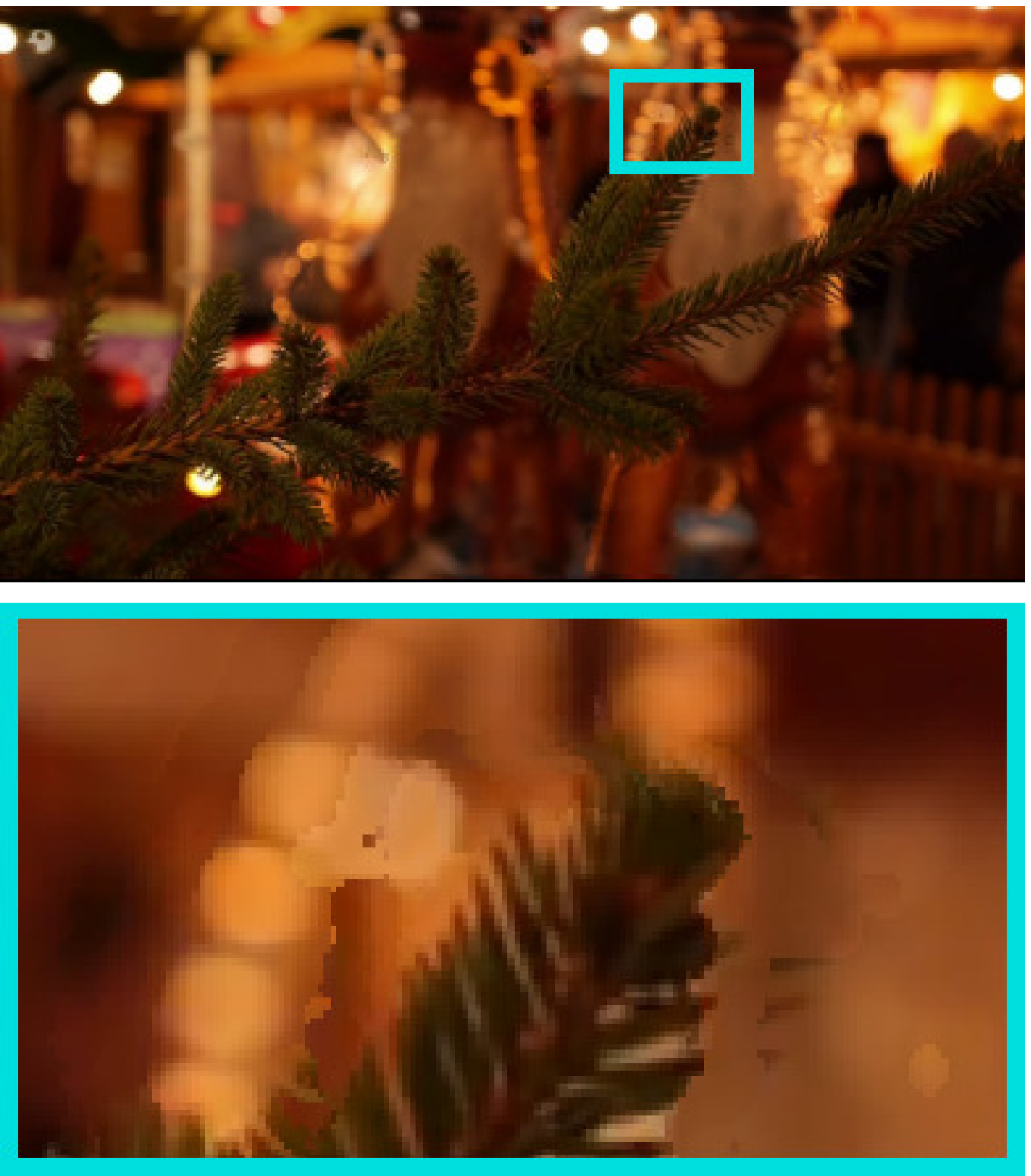}
        \vspace{-0.1cm} \\
    \end{tabularx}
    \begin{tabularx}{\textwidth}{c @{\hspace{0.05cm}} c @{\hspace{0.05cm}} c @{\hspace{0.05cm}} c @{\hspace{0.05cm}} c @{\hspace{0.05cm}} c @{\hspace{0.05cm}} c}
            \includegraphics[width=\itemwidth]{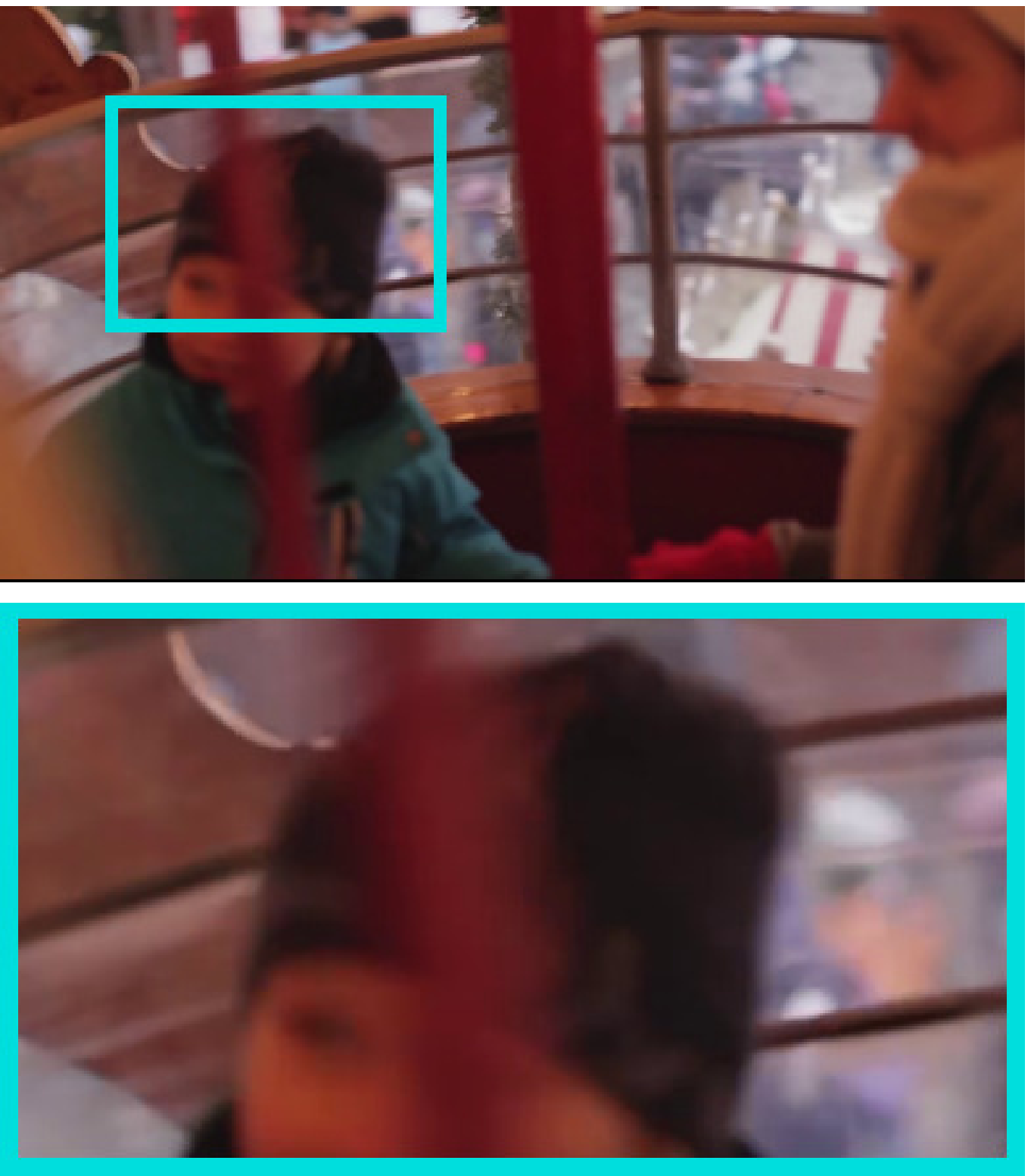}
        &
            \includegraphics[width=\itemwidth]{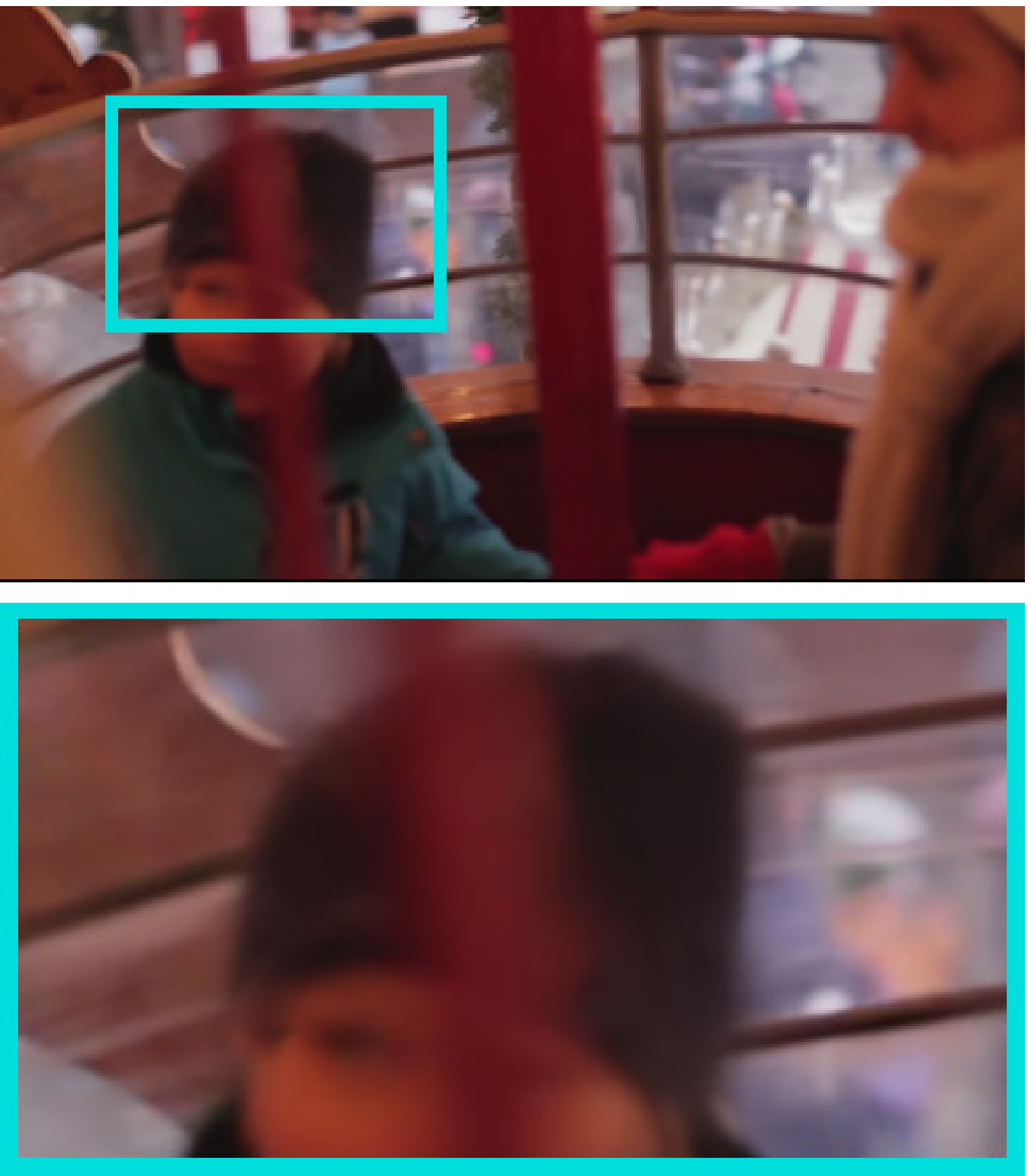}
        &
            \includegraphics[width=\itemwidth]{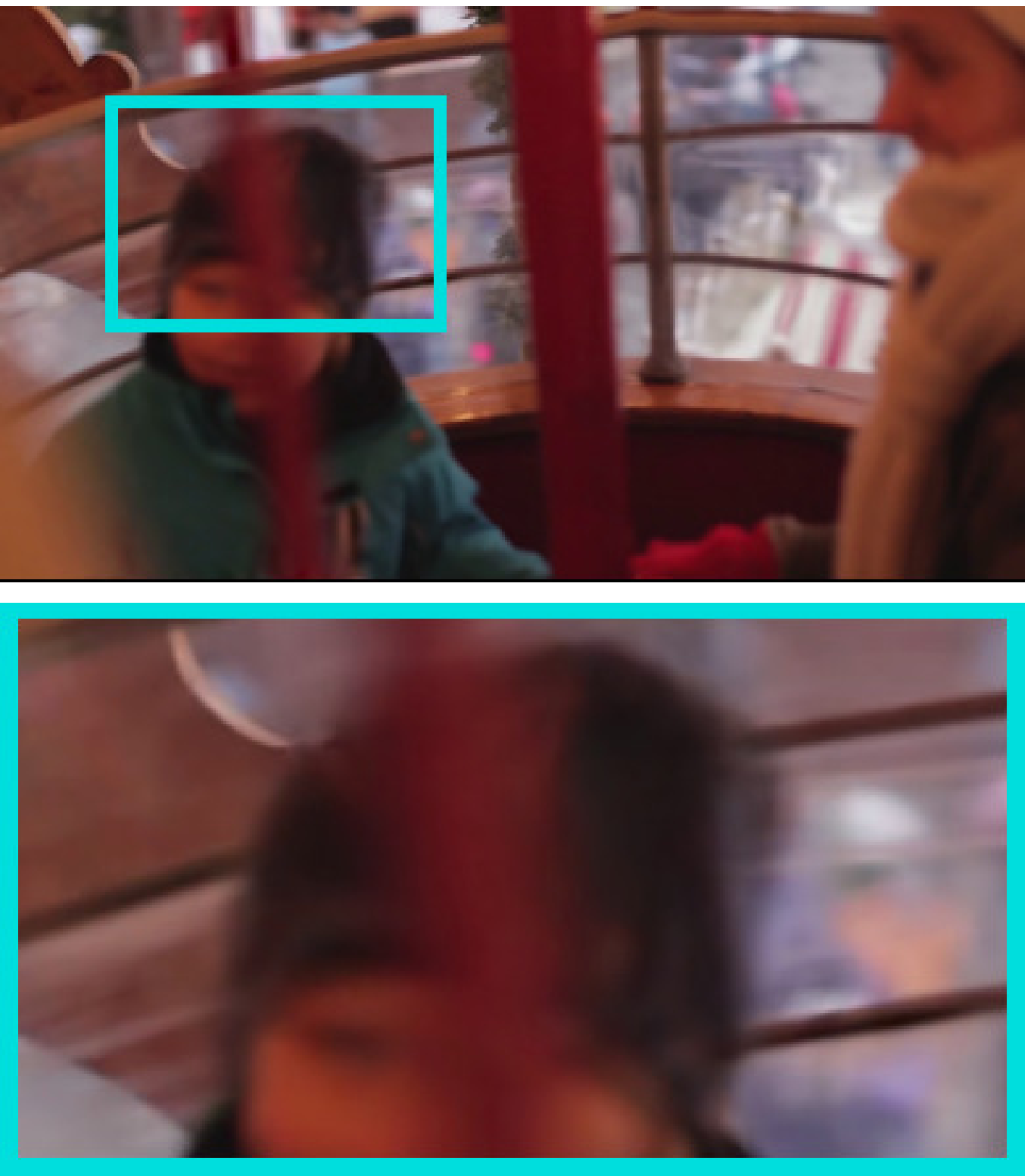}
        &
            \includegraphics[width=\itemwidth]{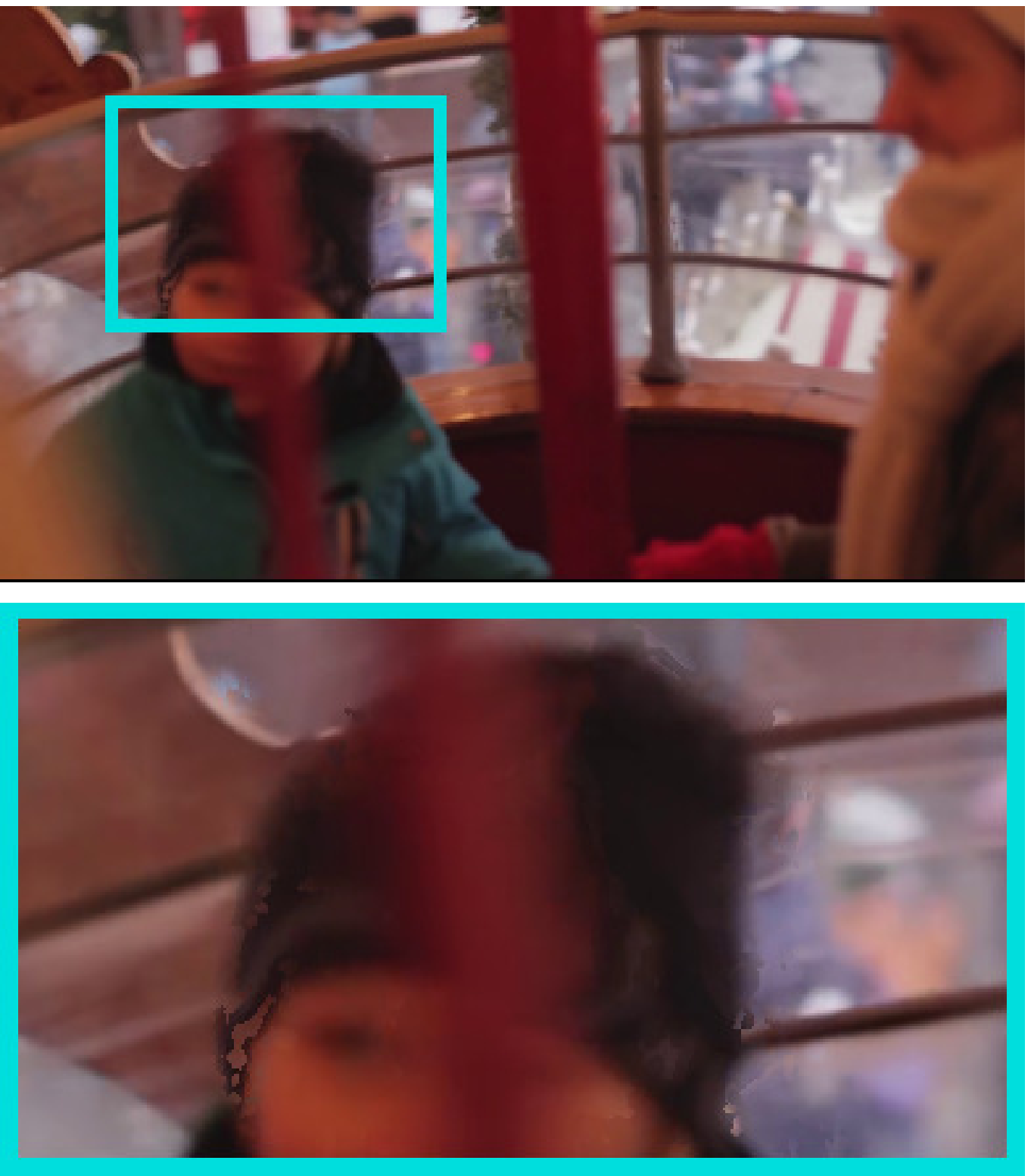}
        &
            \includegraphics[width=\itemwidth]{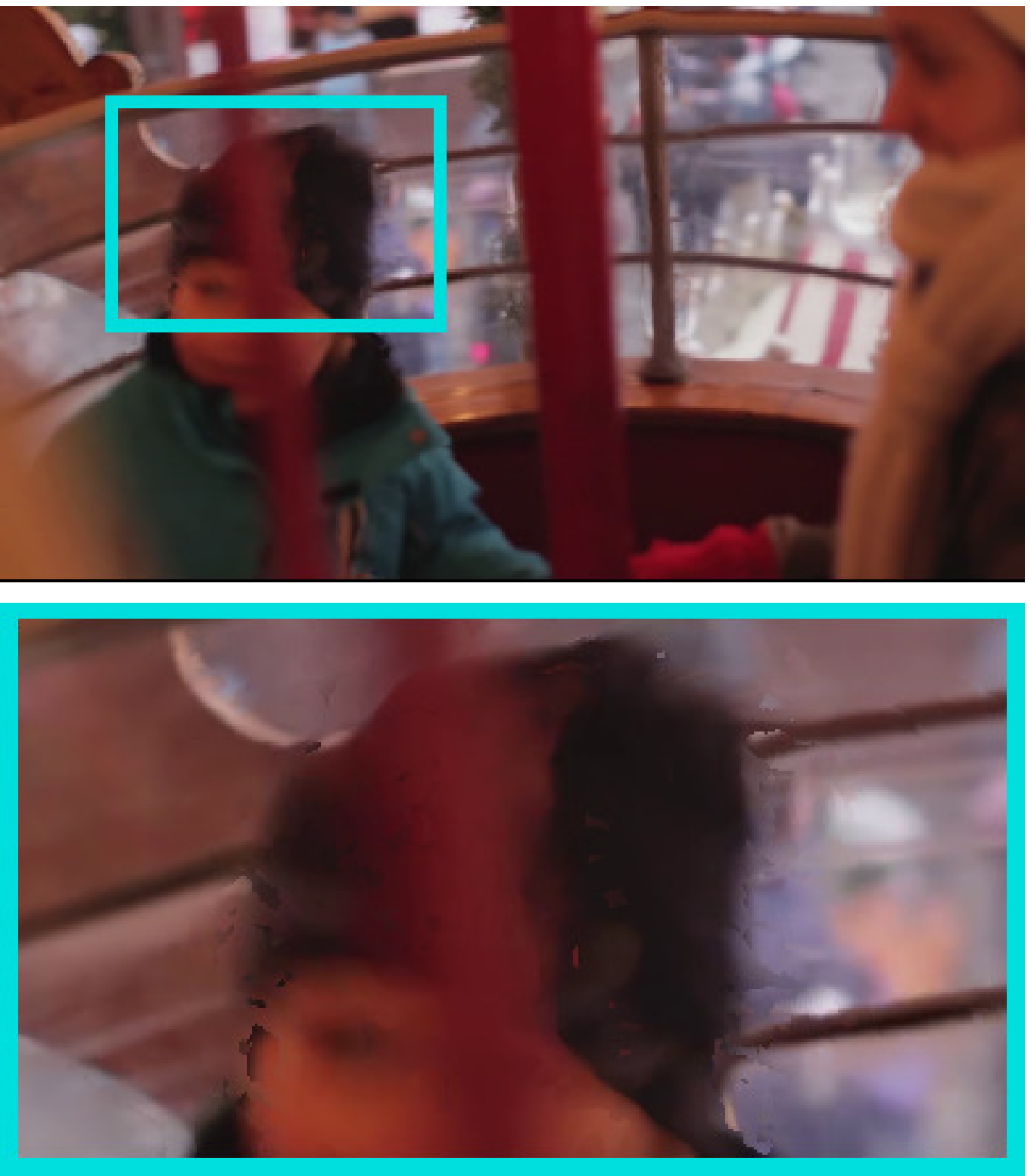}
        &
            \includegraphics[width=\itemwidth]{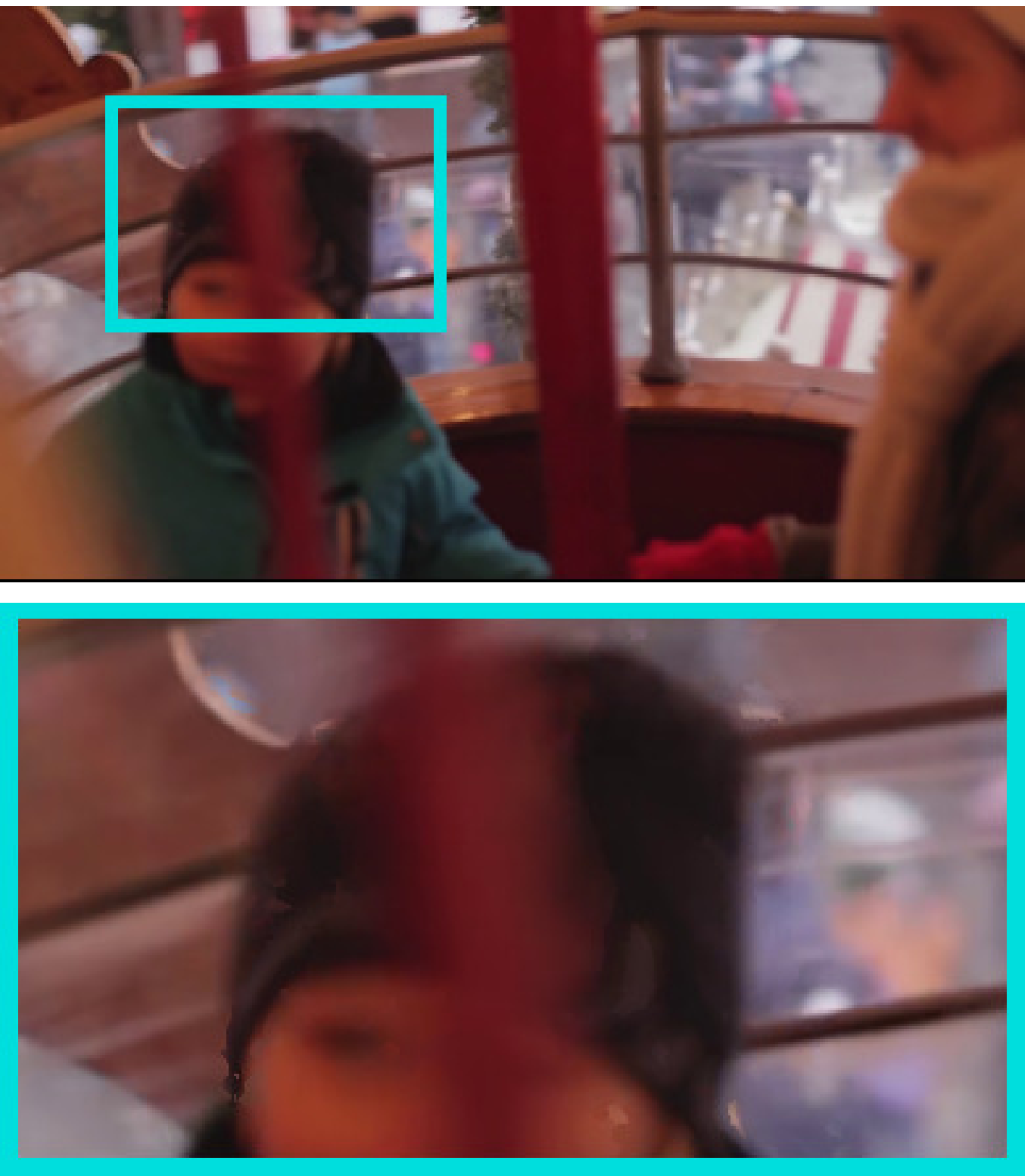}
        &
            \includegraphics[width=\itemwidth]{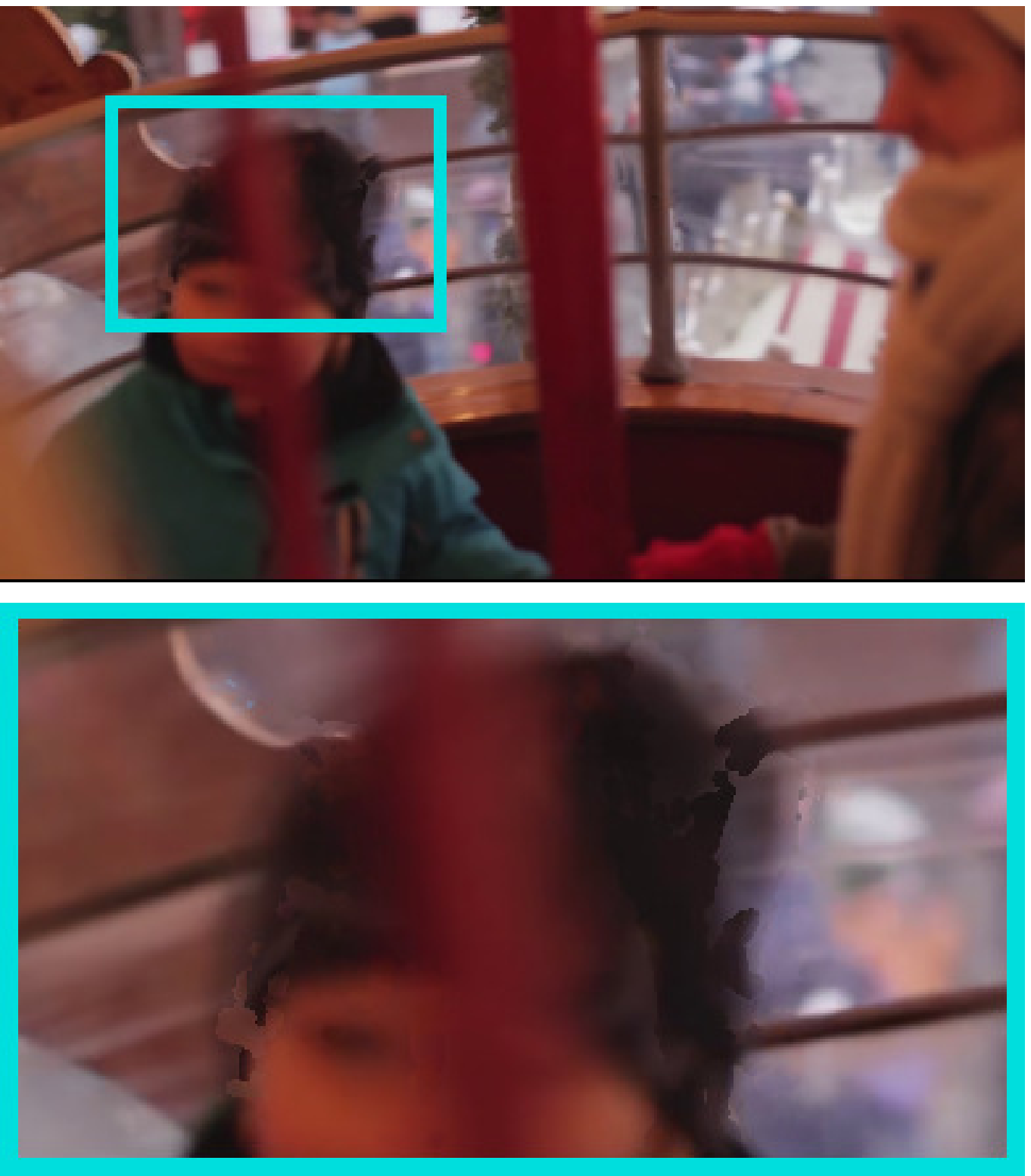}
        \vspace{-0.1cm} \\
            \footnotesize Input frame 1
        &
            \footnotesize Ours
        &
            \footnotesize Meyer~\etal
        &
            \footnotesize DeepFlow2
        &
            \footnotesize FlowNetS
        &
            \footnotesize MDP-Flow2
        &
            \footnotesize Brox~\etal
        \\
    \end{tabularx}\vspace{-0.1in}
    \caption{Qualitative evaluation on blurry videos.}\vspace{-0.15in}
    \label{fig:blurriness}
\end{figure*}

We compare our method to state-of-the-art video frame interpolation methods, including the recent phase-based interpolation method~\cite{Meyer_CVPR_2015} and a few optical flow-based methods. The optical flow algorithms in our experiment include MDP-Flow2~\cite{Xu_PAMI_2012}, which currently produces the lowest interpolation error according to the Middlebury benchmark, the method from Brox \etal~\cite{Brox_ECCV_2004}, as well as two recent deep learning based approaches, namely DeepFlow2~\cite{Weinzaepfel_ICCV_2013} and FlowNetS~\cite{Dosovitskiy_ICCV_2015}. Following recent frame interpolation work~\cite{Meyer_CVPR_2015}, we use the interpolation method from the Middlebury benchmark~\cite{Baker_OTHER_2011} to synthesize the interpolated frame using the optical flow results. Alternatively, other advanced image-based rendering algorithms~\cite{Zitnick_TOG_2004} can also be used. For the two deep learning-based optical flow methods, we directly use the trained models from the author websites.

\subsection{Comparisons}

We evaluate our method quantitatively on the Middlebury optical flow benchmark~\cite{Baker_OTHER_2011}. As reported in Table~\ref{tbl:middlebury}, our method performs very well on the four examples with real-world scenes. Among the over 100 methods reported in the Middlebury benchmark, our method achieves the best on Evergreen and Basketball, 2nd best on Dumptruck, and 3rd best on Backyard. Our method does not work as well on the other four examples that are either synthetic or of lab scenes, partially because we train our network on videos with real-world scenes. Qualitatively, we find that our method can often create results in challenging regions that are visually more appealing than  state-of-the-art methods. 

\noindent\textbf{Blur.} Figure~\ref{fig:blurriness} shows two examples where the input videos suffer from out-of-focus blur (top) and motion blur (bottom). Blurry regions are often challenging for optical flow estimation; thus these regions in the interpolated results suffer from noticeable artifacts. Both our method and the phase-based method from Meyer~\etal~\cite{Meyer_CVPR_2015} can handle blurry regions better while our method produces sharper images, especially in regions with large motion, such as the right side of the hat in the bottom example.

\begin{figure*}\centering
    \setlength{\tabcolsep}{0.0cm}
    \setlength{\itemwidth}{2.45cm}

    \begin{tabularx}{\textwidth}{c @{\hspace{0.05cm}} c @{\hspace{0.05cm}} c @{\hspace{0.05cm}} c @{\hspace{0.05cm}} c @{\hspace{0.05cm}} c @{\hspace{0.05cm}} c}
            \begin{minipage}[b]{\itemwidth}
                \includegraphics[width=\itemwidth]{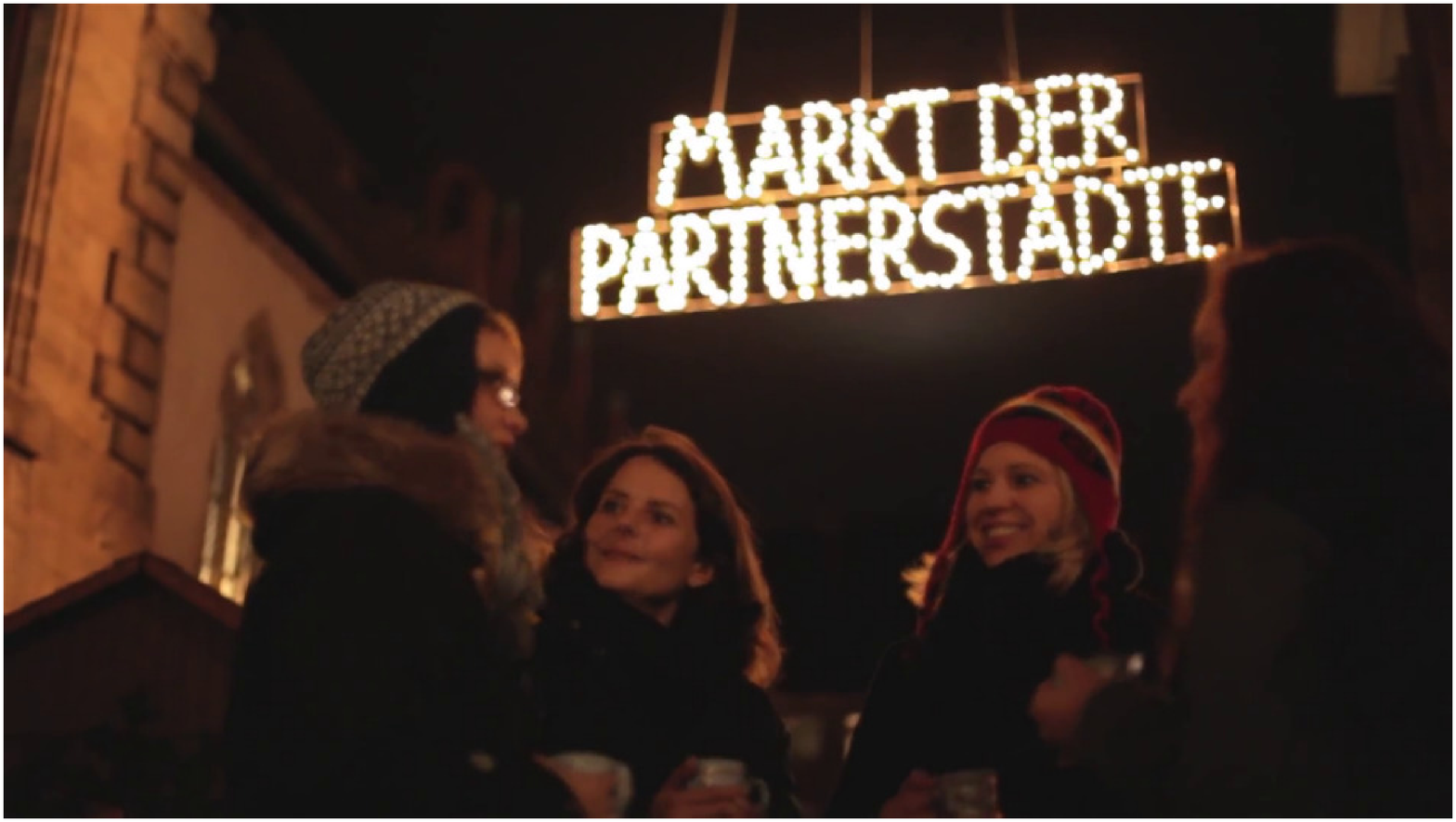}
                \\[0.01cm]
                \includegraphics[width=\itemwidth]{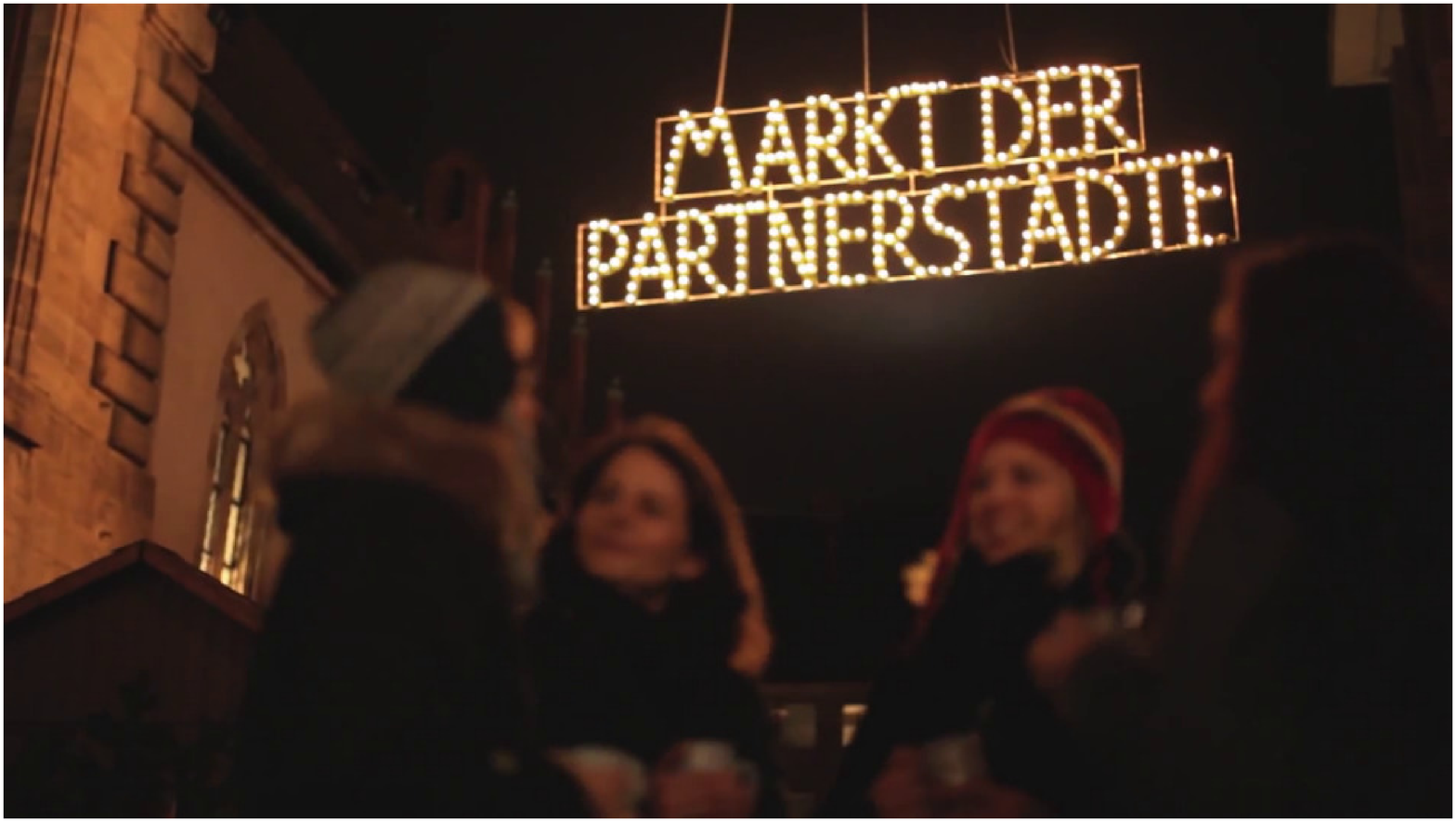}
            \end{minipage}
        &
            \includegraphics[width=\itemwidth]{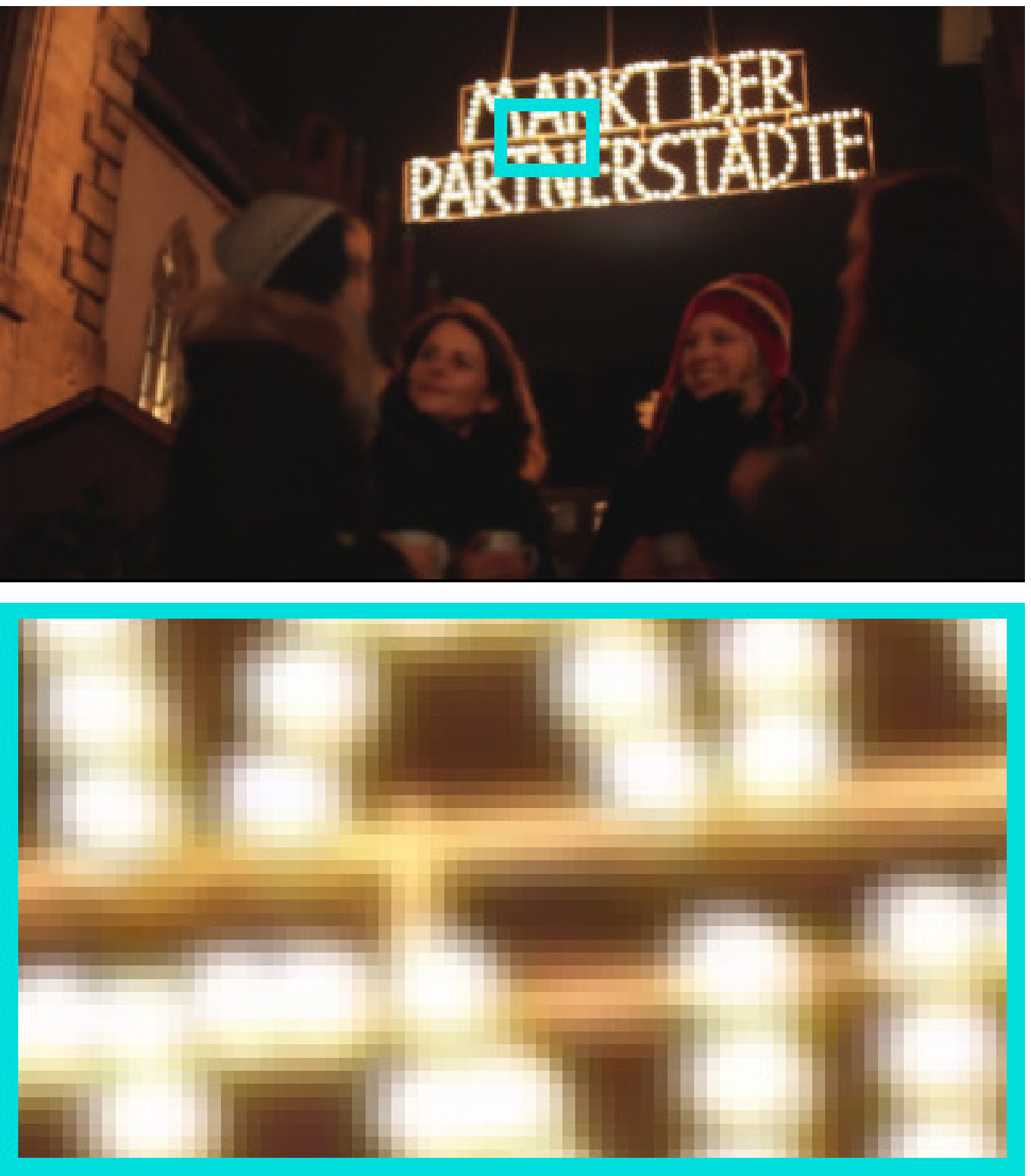}
        &
            \includegraphics[width=\itemwidth]{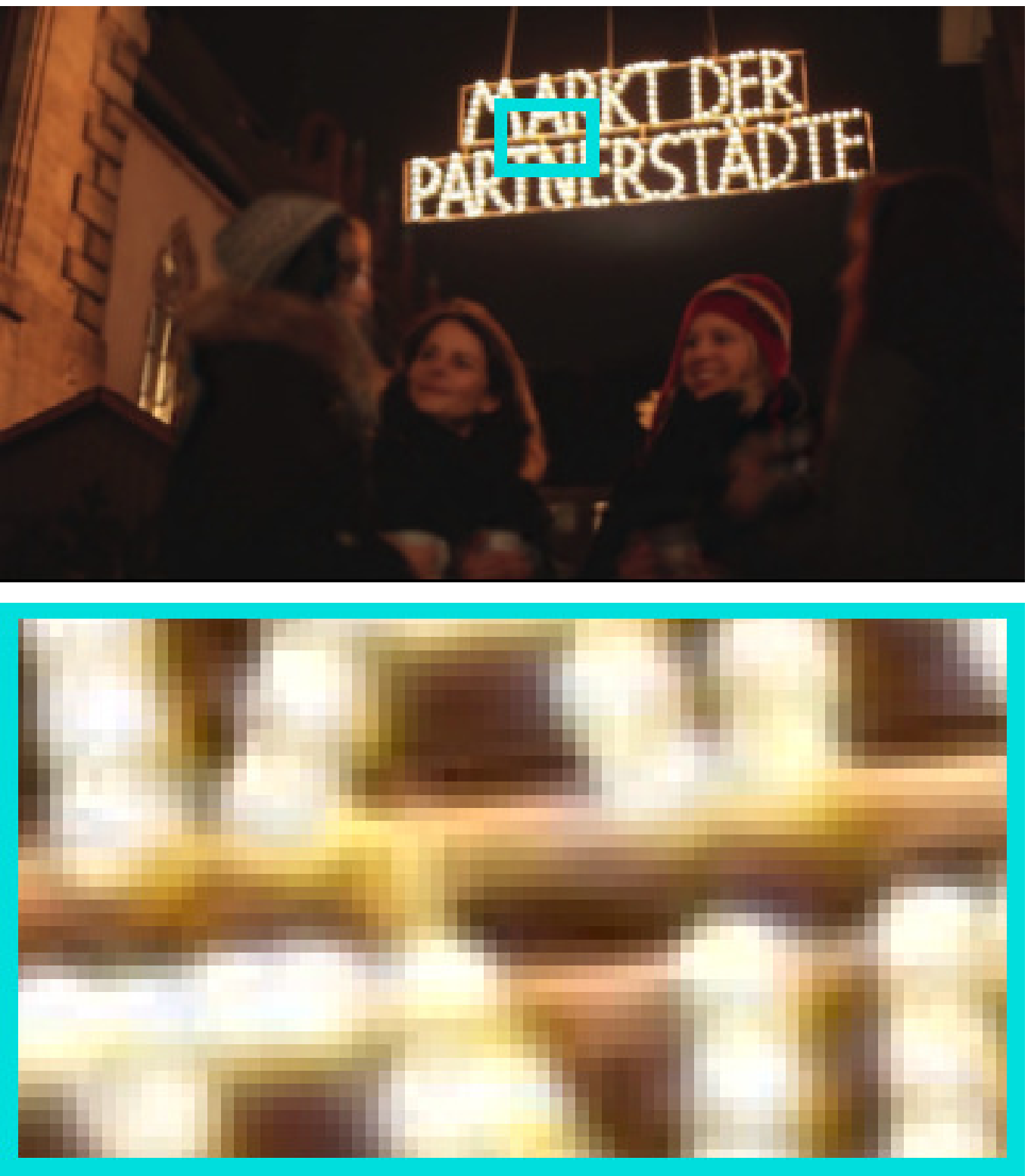}
        &
            \includegraphics[width=\itemwidth]{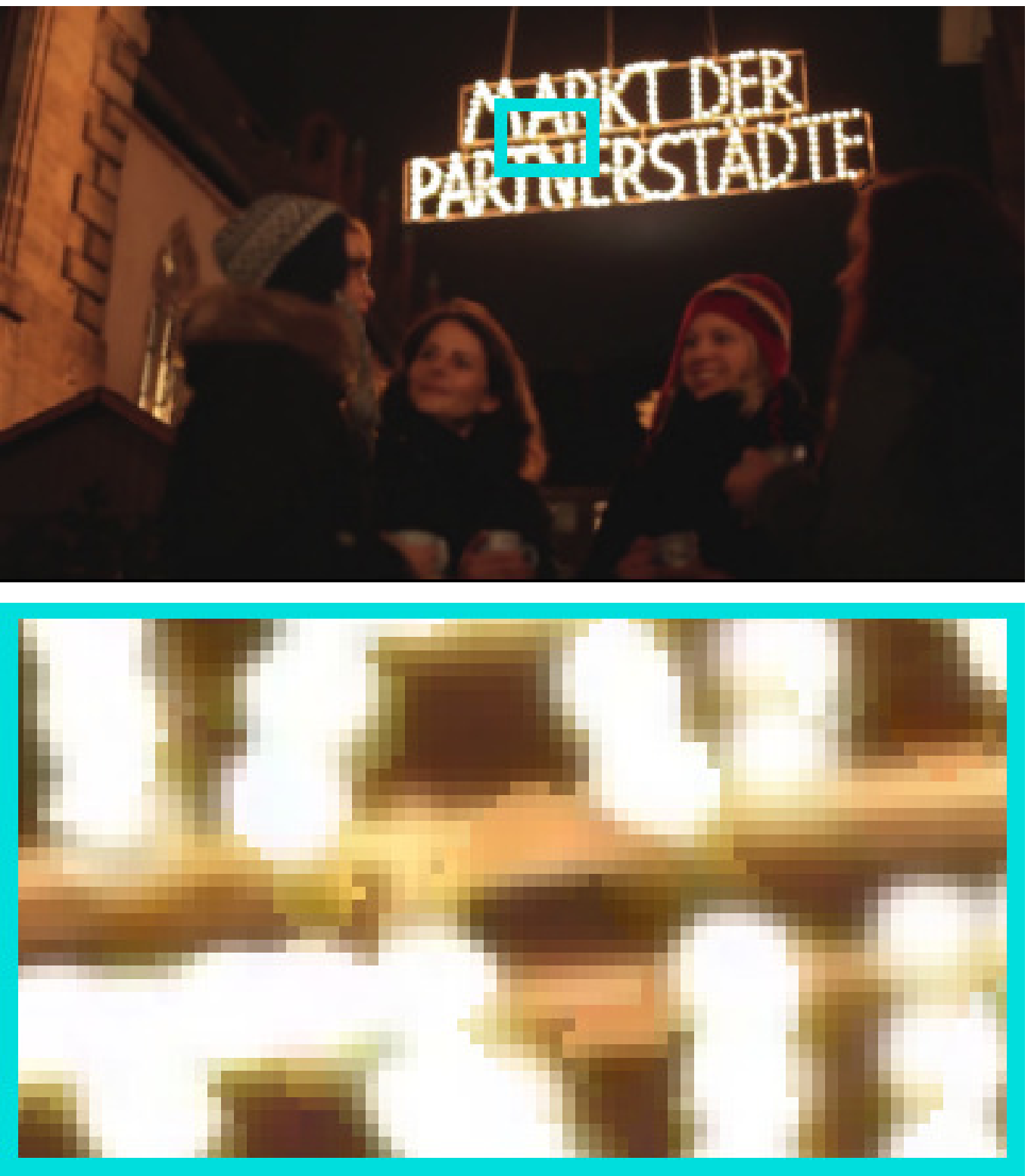}
        &
            \includegraphics[width=\itemwidth]{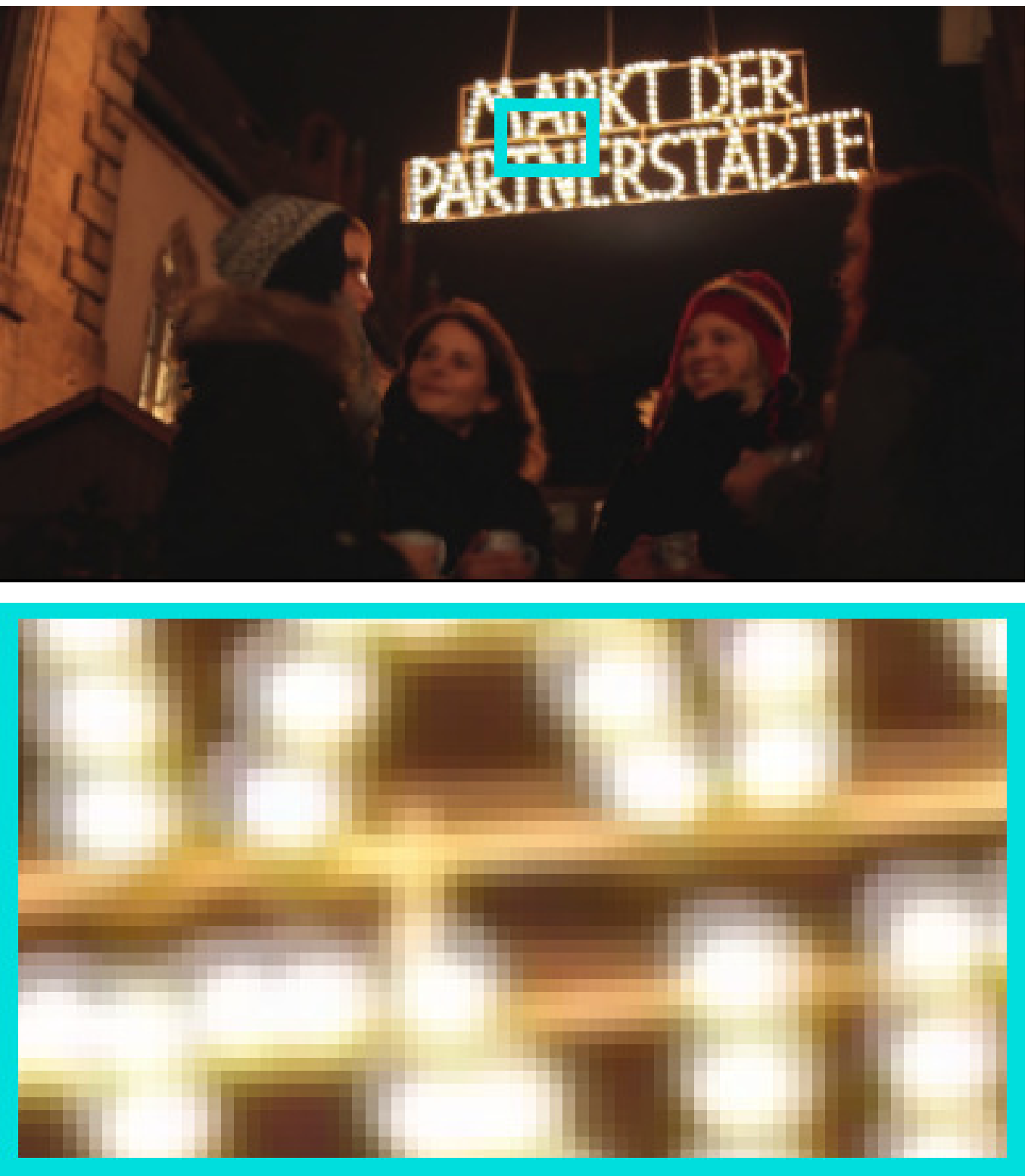}
        &
            \includegraphics[width=\itemwidth]{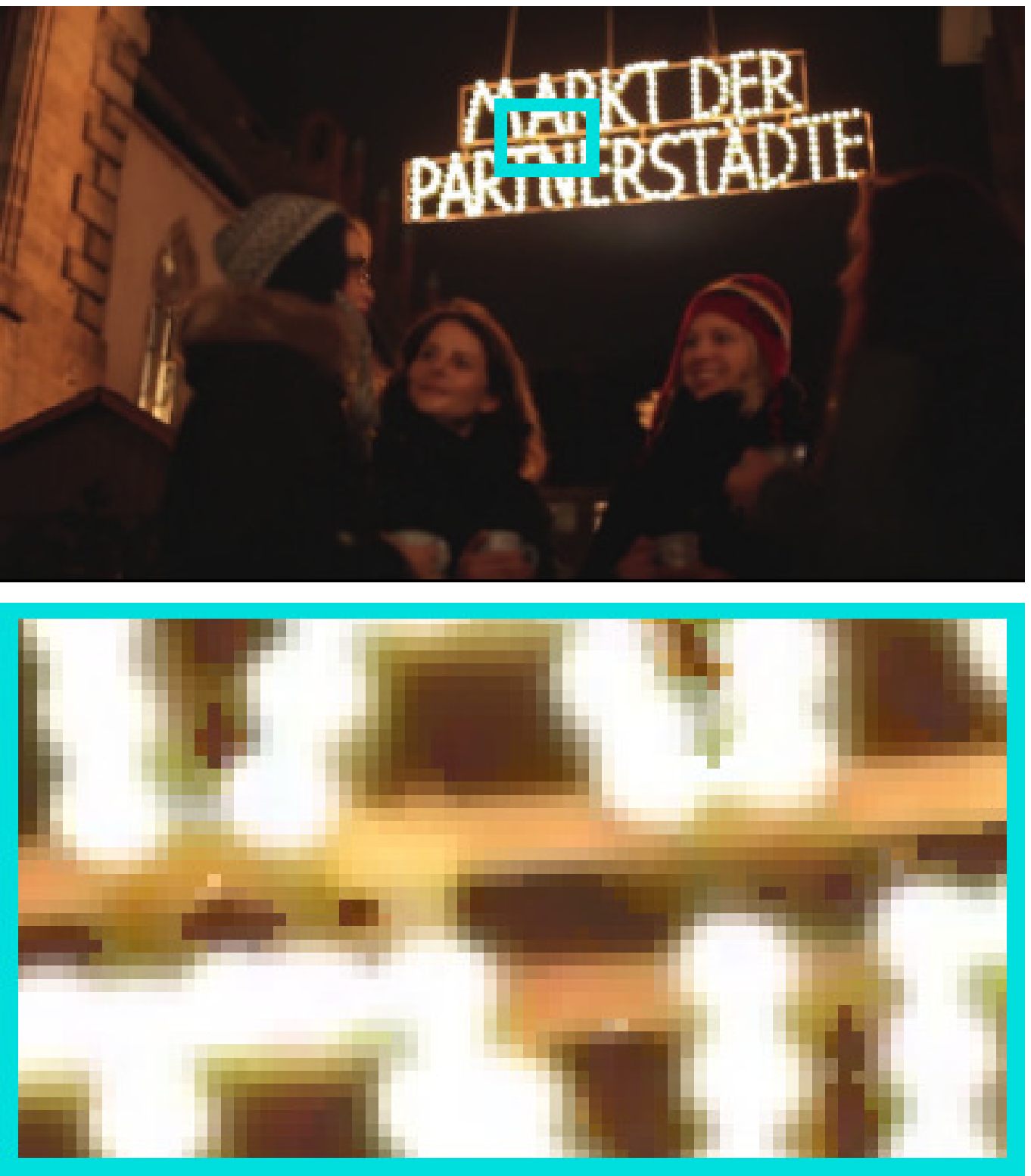}
        &
            \includegraphics[width=\itemwidth]{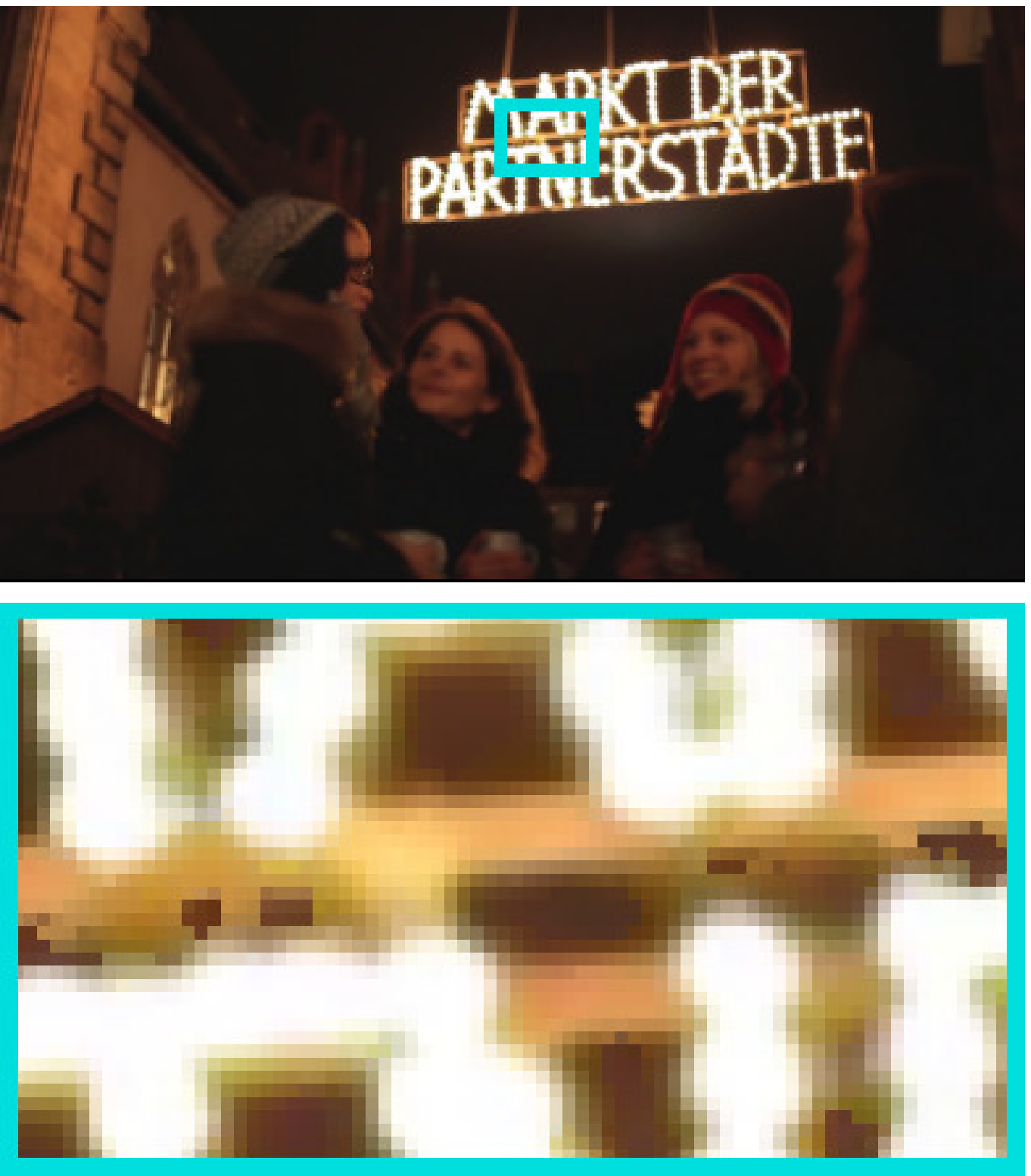}
        \vspace{-0.1cm} \\
            \footnotesize Input frames
        &
            \footnotesize Ours
        &
            \footnotesize Meyer~\etal
        &
            \footnotesize DeepFlow2
        &
            \footnotesize FlowNetS
        &
            \footnotesize MDP-Flow2
        &
            \footnotesize Brox~\etal
        \\
    \end{tabularx}\vspace{-0.1in}
    \caption{Qualitative evaluation on video with abrupt brightness change.}\vspace{-0.1in}
    \label{fig:brightness}
\end{figure*}

\noindent\textbf{Abrupt brightness change.} As shown in Figure~\ref{fig:brightness}, abrupt brightness change violates the brightness consistency assumption and compromises optical flow estimation, causing artifacts in frame interpolation. For this example, our method and the phase-based method generate more visually appealing interpolation results than flow-based methods.

\begin{table}\centering\scriptsize
    \begin{tabularx}{\columnwidth}{X c @{\hspace{0.15cm}} c @{\hspace{0.15cm}} c @{\hspace{0.15cm}} c @{\hspace{0.15cm}} c @{\hspace{0.15cm}} c @{\hspace{0.15cm}} c @{\hspace{0.15cm}} c}
        \toprule
            & Mequ. & Schef. & Urban & Teddy & Backy. & Baske. & Dumpt. & Everg.
        \\ \midrule
            Ours & $3.57$ & $4.34$ & $5.00$ & $6.91$ & $\bm{10.2}$ & $\bm{5.33}$ & $\bm{7.30}$ & $\bm{6.94}$
        \\
            DeepFlow2 & $2.99$ & $3.88$ & $\bm{3.62}$ & $5.38$ & $11.0$ & $5.83$ & $7.60$ & $7.82$
        \\
            FlowNetS & $3.07$ & $4.57$ & $4.01$ & $5.55$ & $11.3$ & $5.99$ & $8.63$ & $7.70$
        \\
            MDP-Flow2 & $\bm{2.89}$ & $\bm{3.47}$ & $3.66$ & $\bm{5.20}$ & $\bm{10.2}$ & $6.13$ & $7.36$ & $7.75$
        \\
            Brox~\etal & $3.08$ & $3.83$ & $3.93$ & $5.32$ & $10.6$ & $6.60$ & $8.61$ & $7.43$
        \\ \bottomrule
    \end{tabularx}\vspace{-0.1in}
    \normalsize\caption{Evaluation on the Middlebury testing set (average interpolation error).}\vspace{-0.2in}
    \label{tbl:middlebury}
\end{table}

\begin{figure*}\centering
    \setlength{\tabcolsep}{0.0cm}
    \setlength{\itemwidth}{2.45cm}

    \begin{tabularx}{\textwidth}{c @{\hspace{0.05cm}} c @{\hspace{0.05cm}} c @{\hspace{0.05cm}} c @{\hspace{0.05cm}} c @{\hspace{0.05cm}} c @{\hspace{0.05cm}} c}
            \includegraphics[width=\itemwidth]{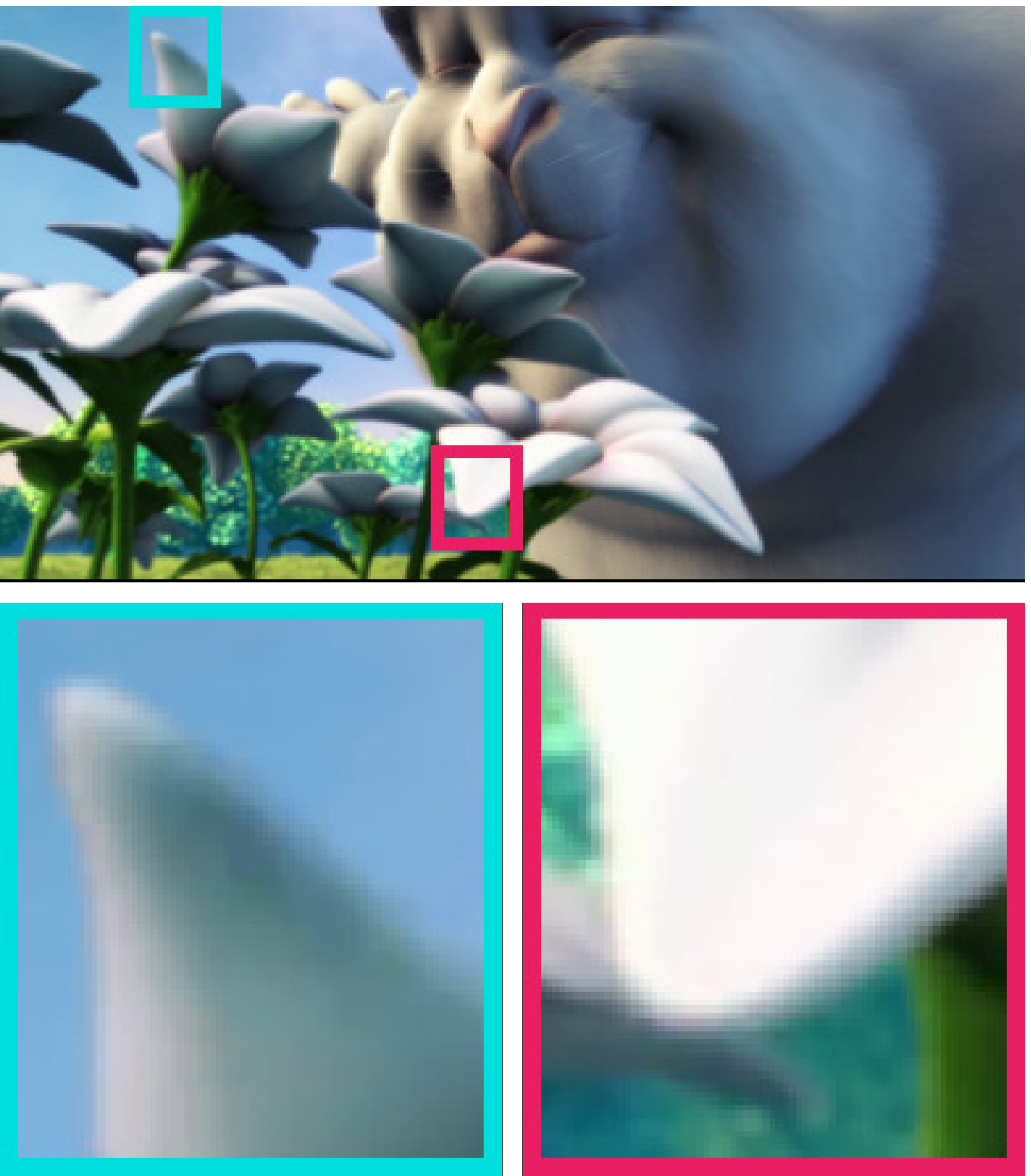}
        &
            \includegraphics[width=\itemwidth]{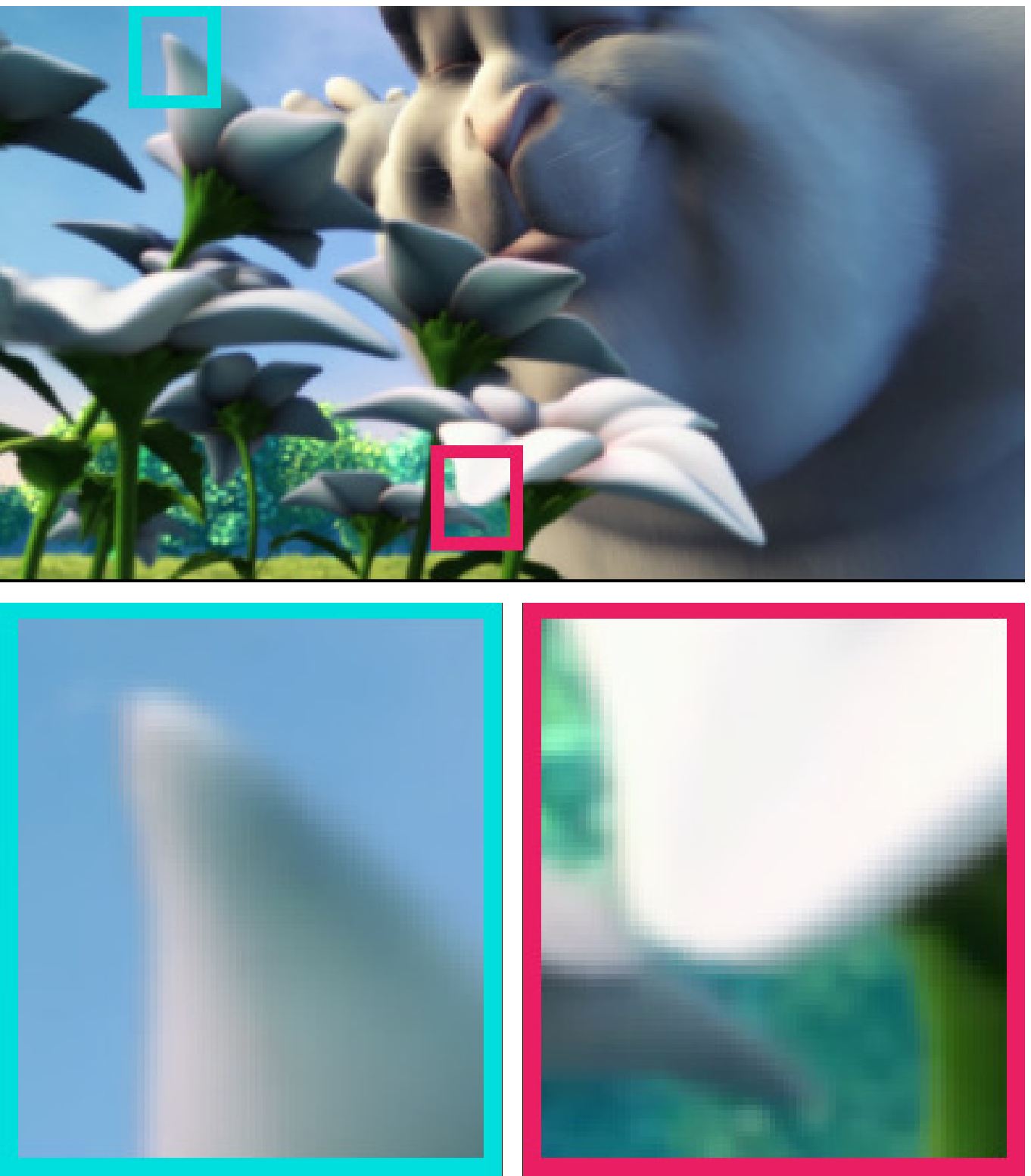}
        &
            \includegraphics[width=\itemwidth]{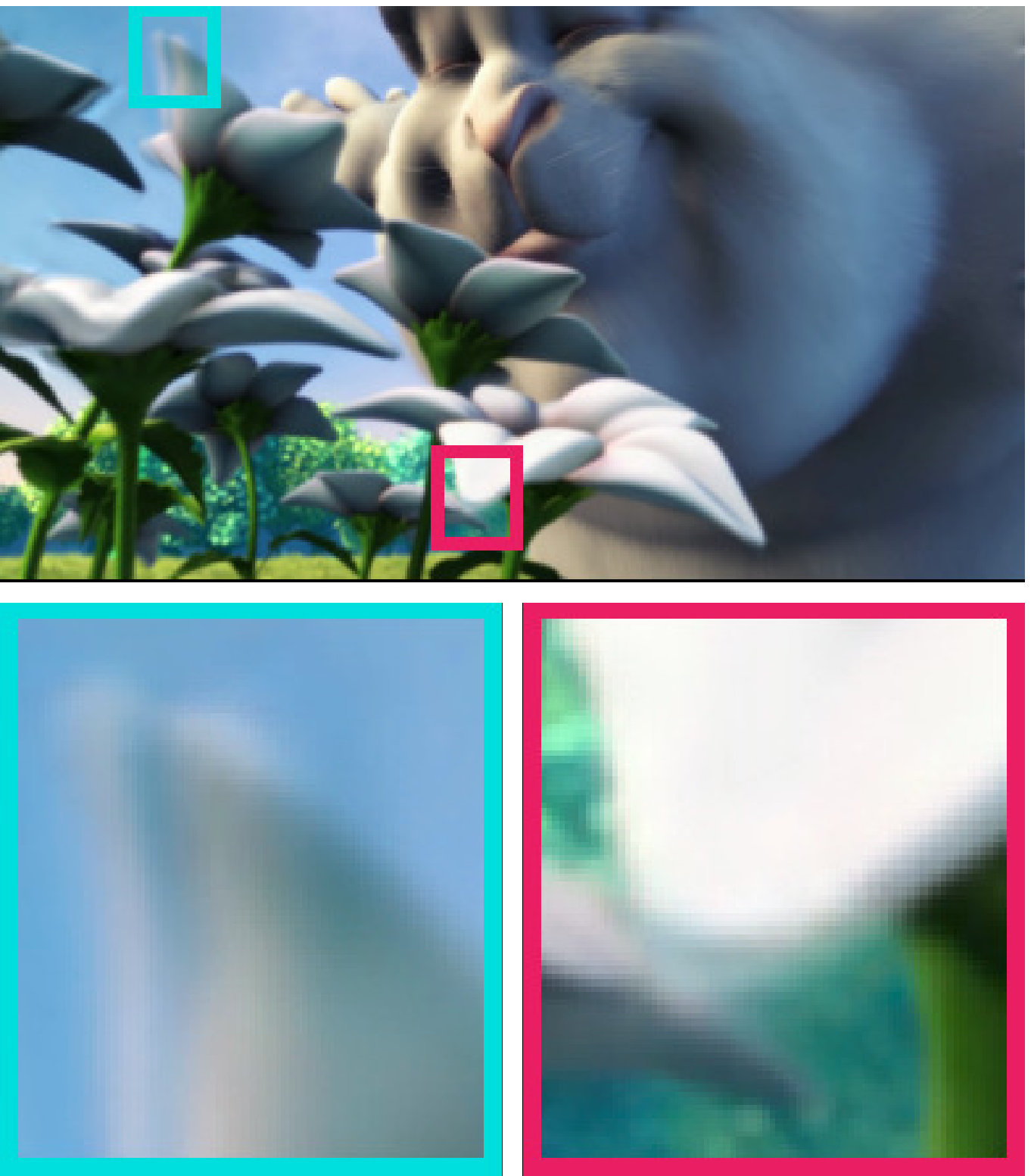}
        &
            \includegraphics[width=\itemwidth]{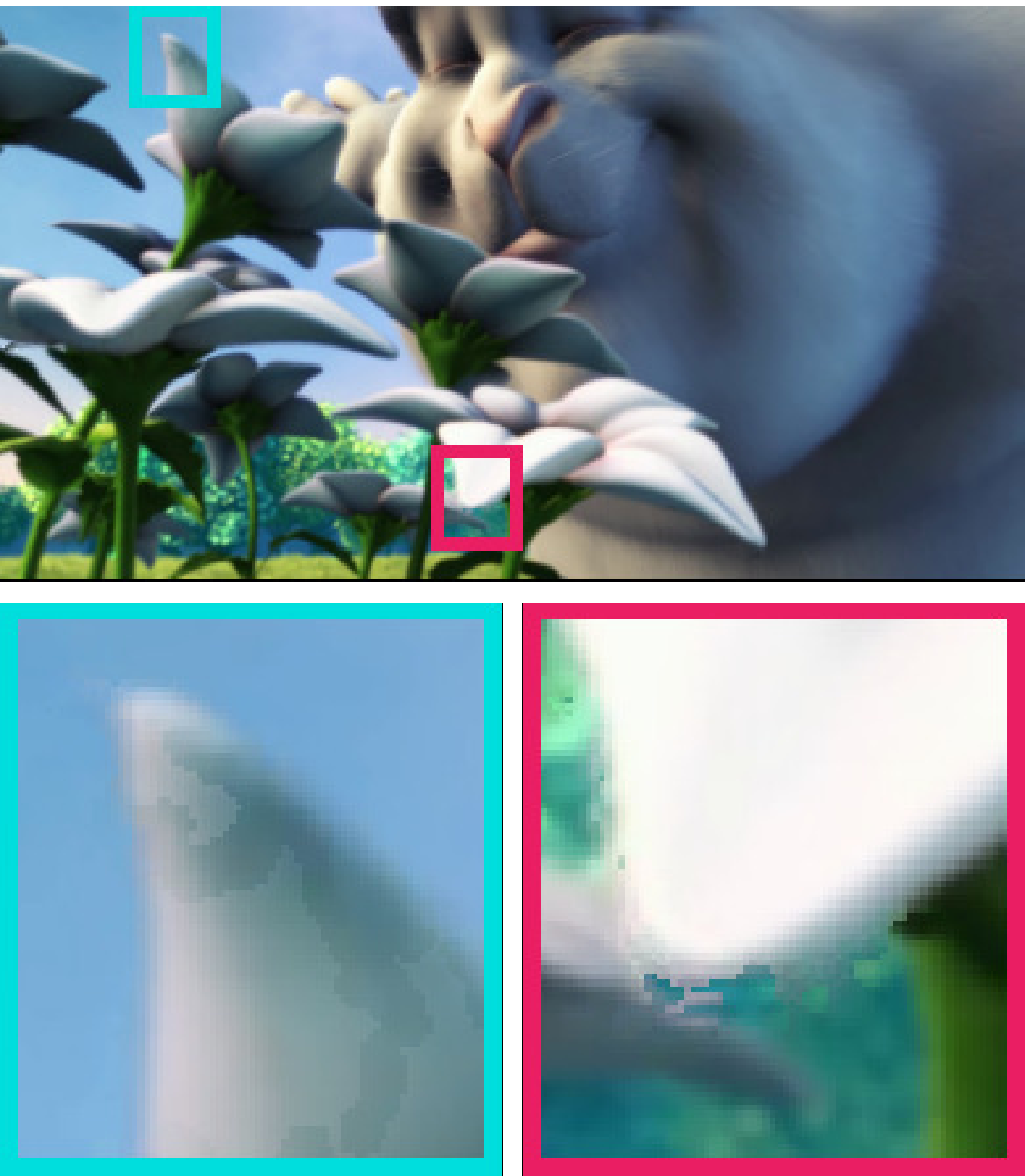}
        &
            \includegraphics[width=\itemwidth]{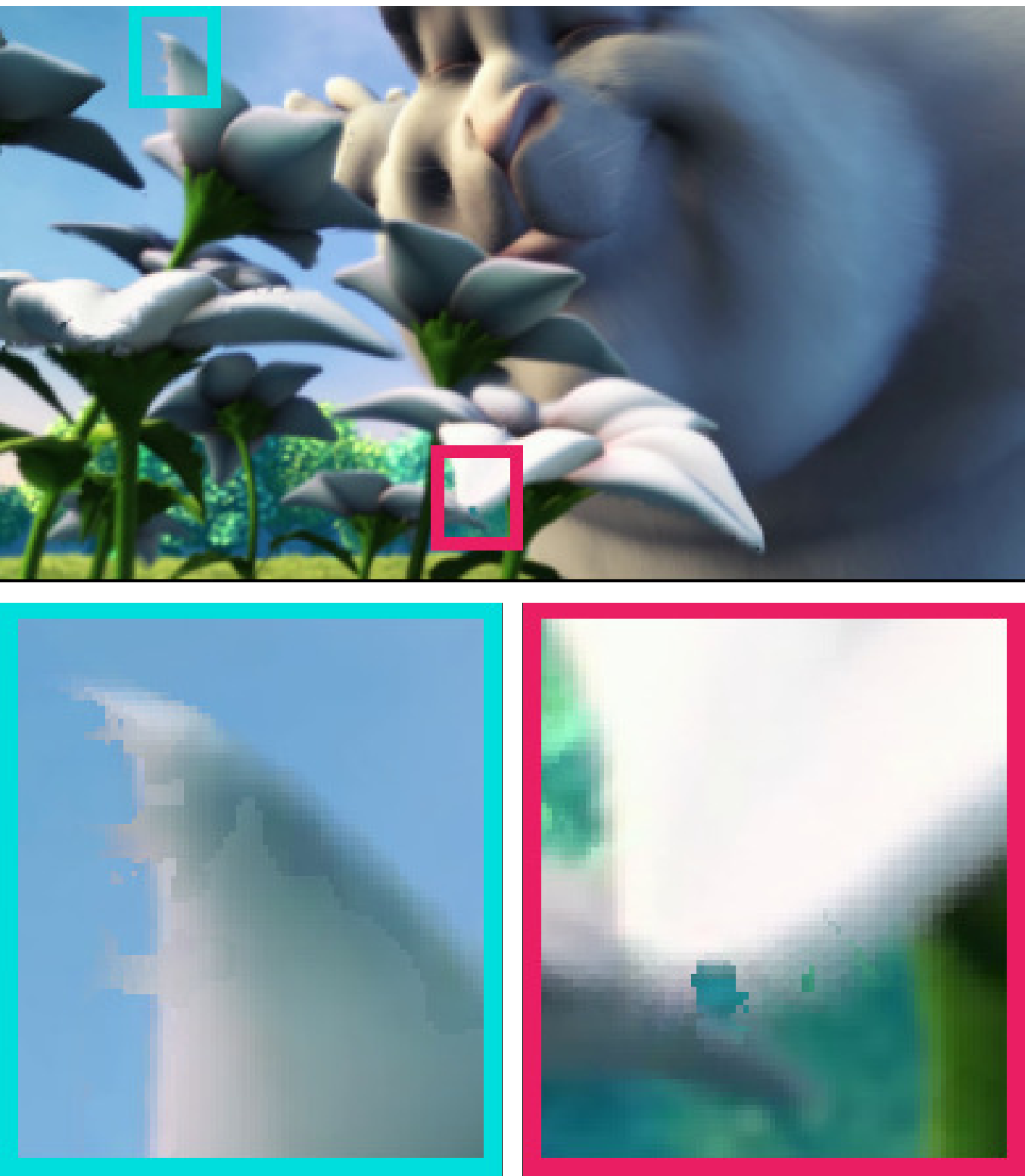}
        &
            \includegraphics[width=\itemwidth]{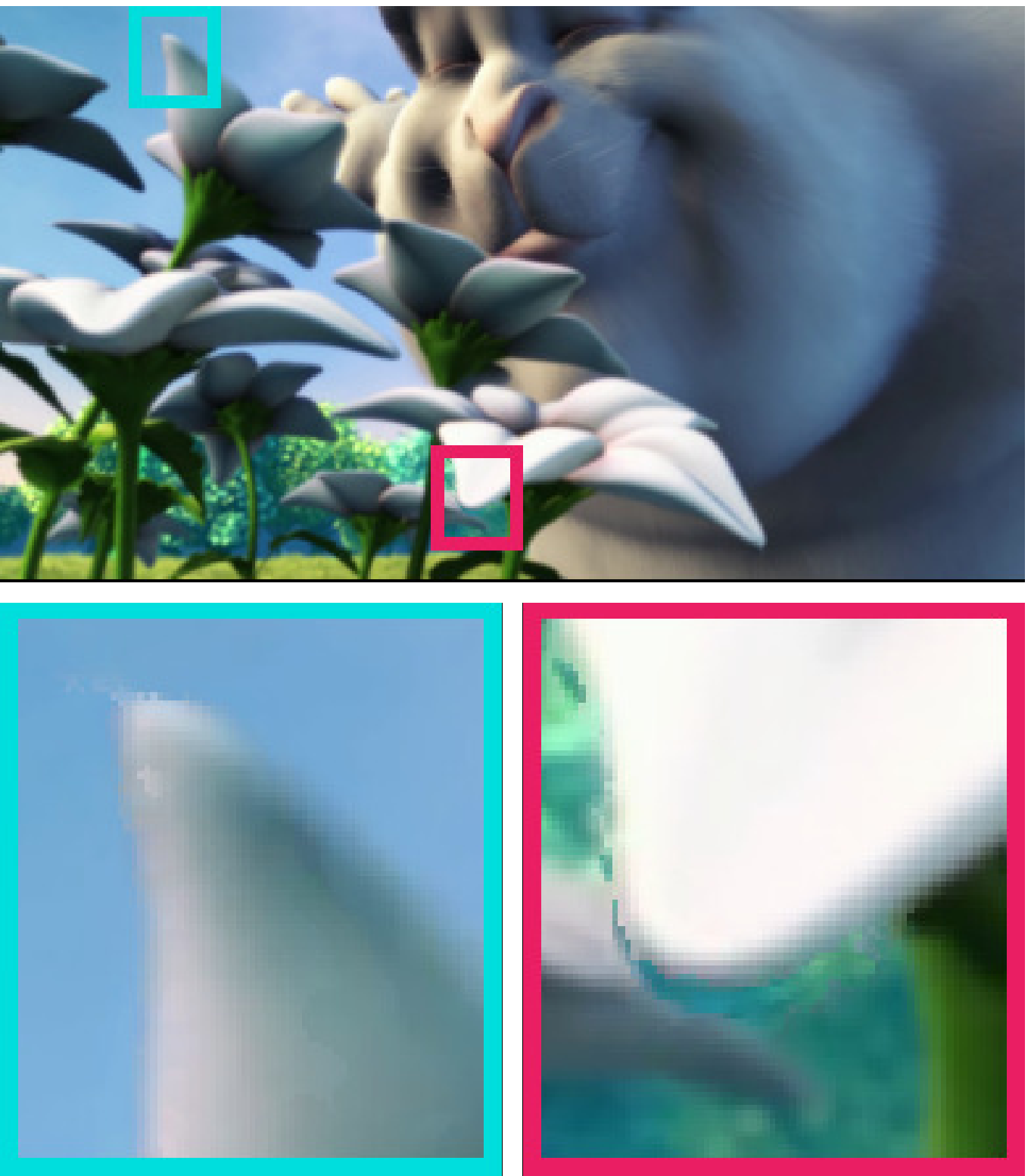}
        &
            \includegraphics[width=\itemwidth]{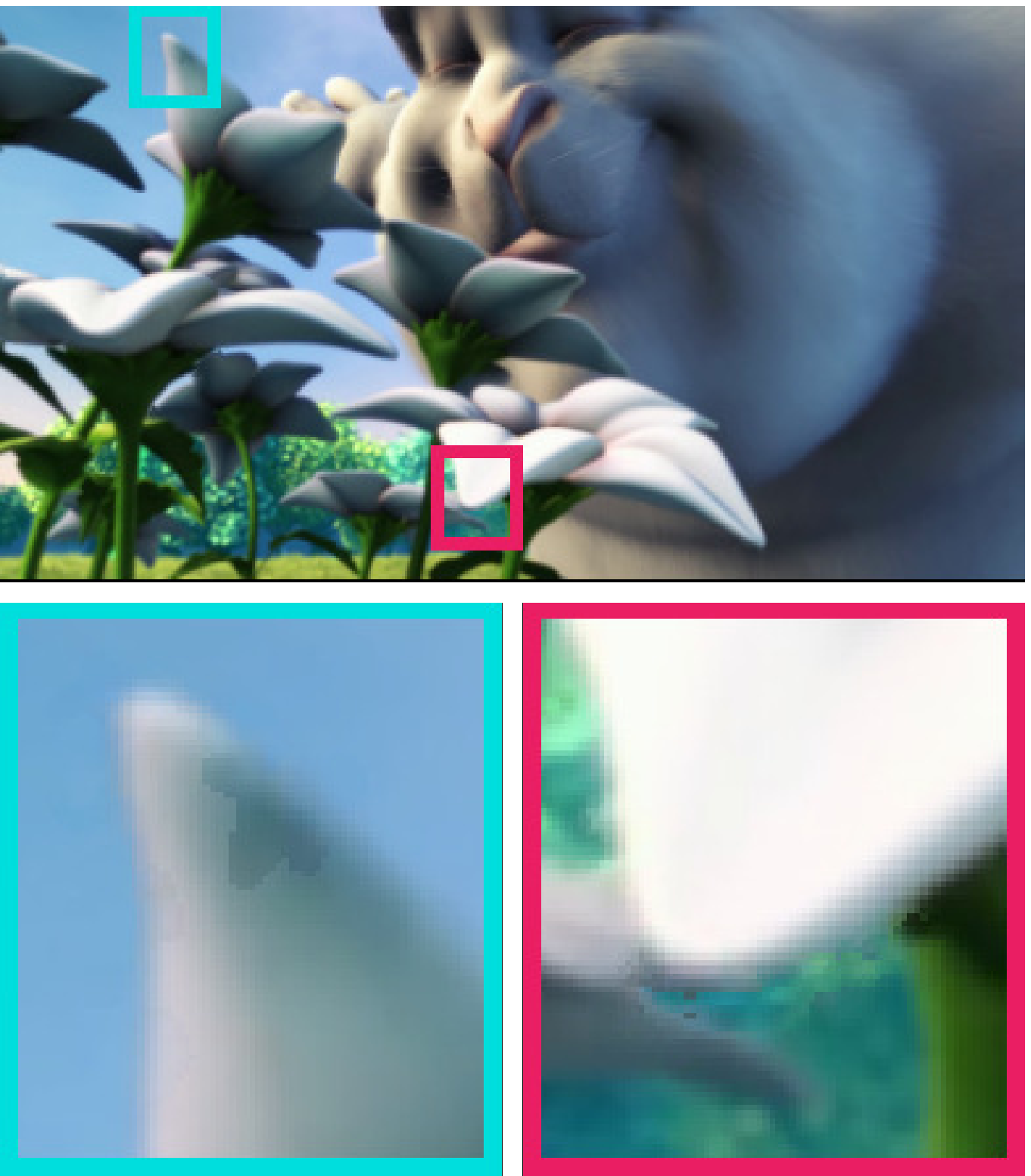}
        \vspace{-0.1cm} \\
    \end{tabularx}
    \begin{tabularx}{\textwidth}{c @{\hspace{0.05cm}} c @{\hspace{0.05cm}} c @{\hspace{0.05cm}} c @{\hspace{0.05cm}} c @{\hspace{0.05cm}} c @{\hspace{0.05cm}} c}
            \includegraphics[width=\itemwidth]{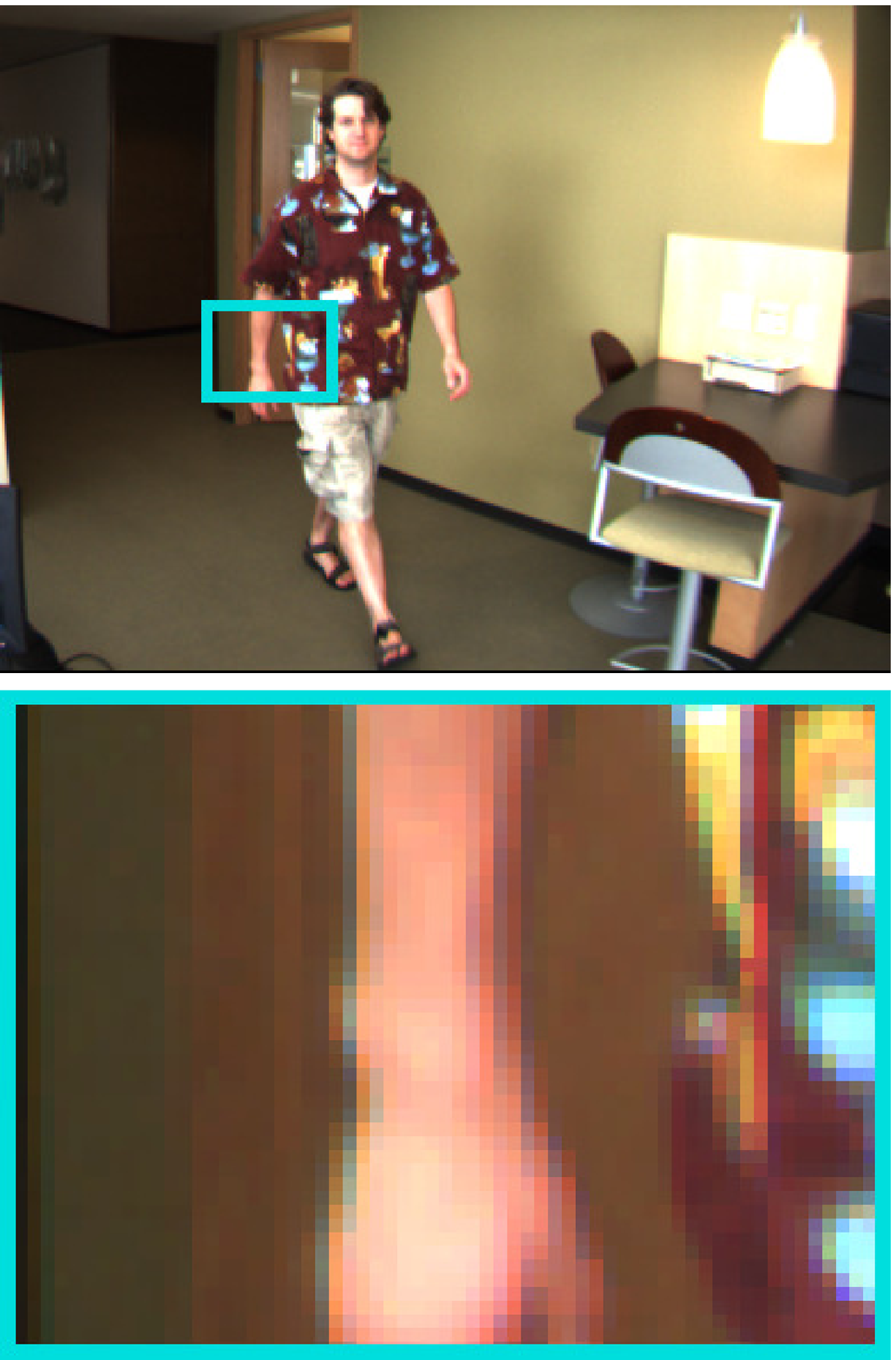}
        &
            \includegraphics[width=\itemwidth]{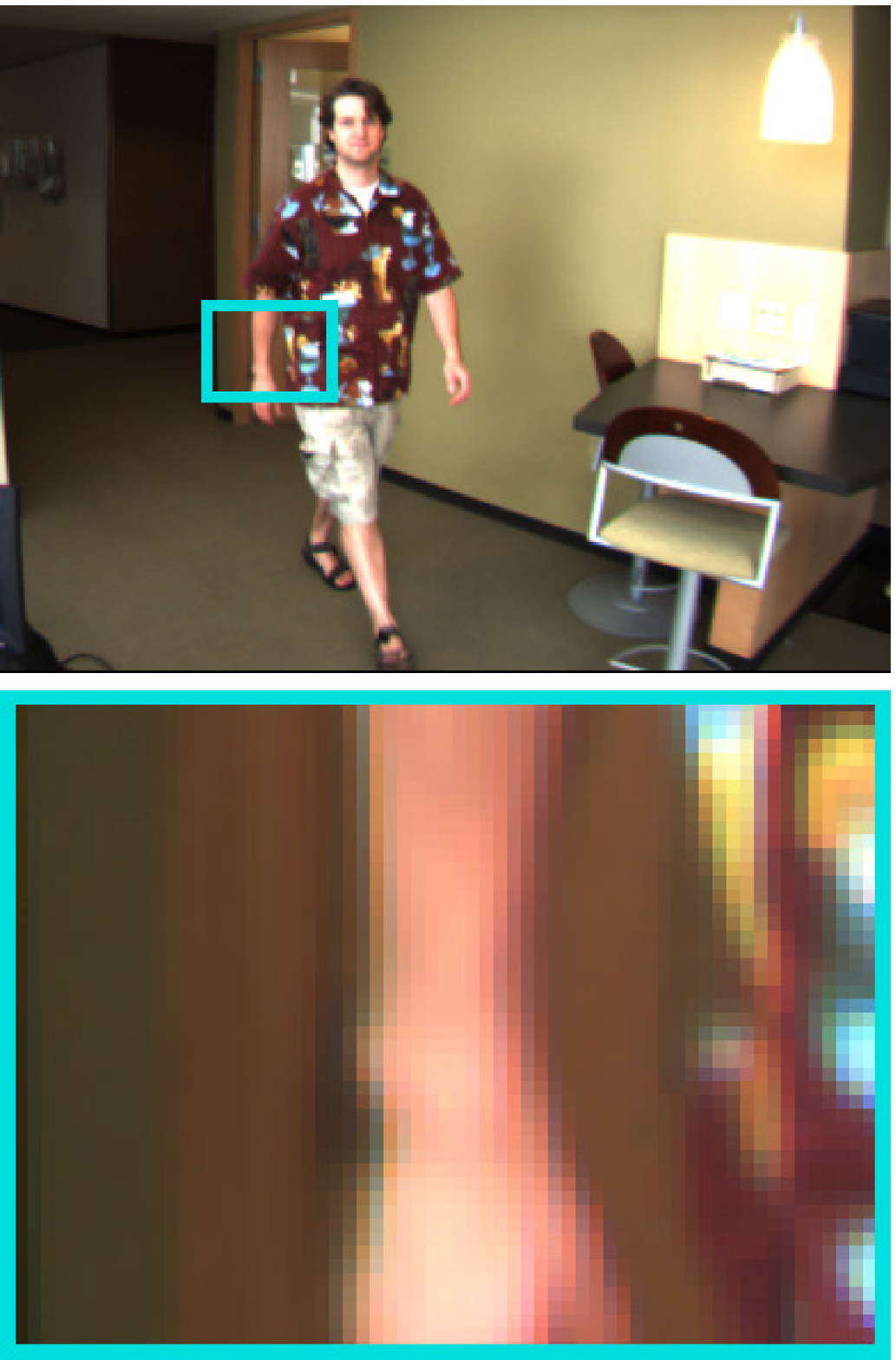}
        &
            \includegraphics[width=\itemwidth]{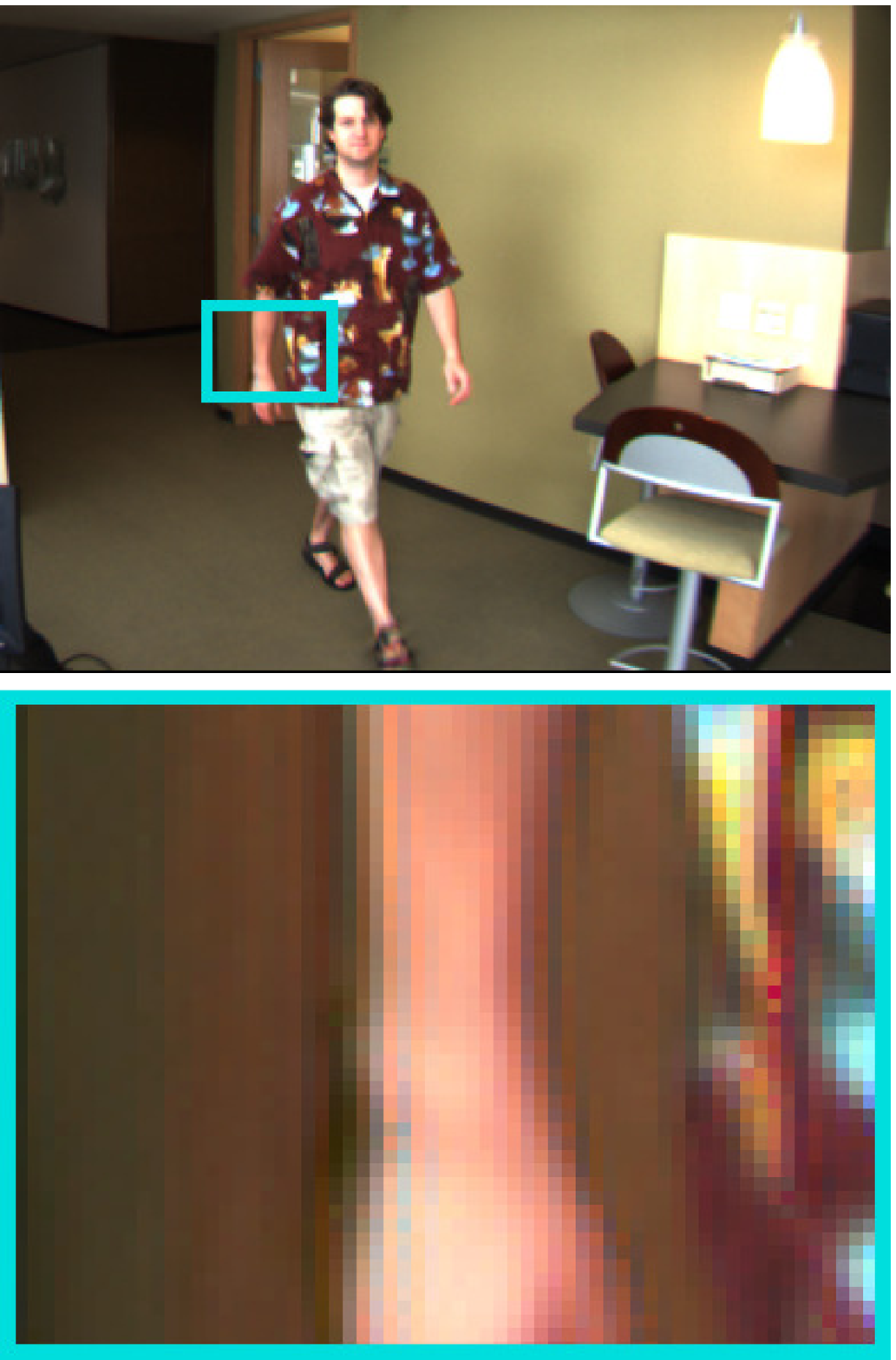}
        &
            \includegraphics[width=\itemwidth]{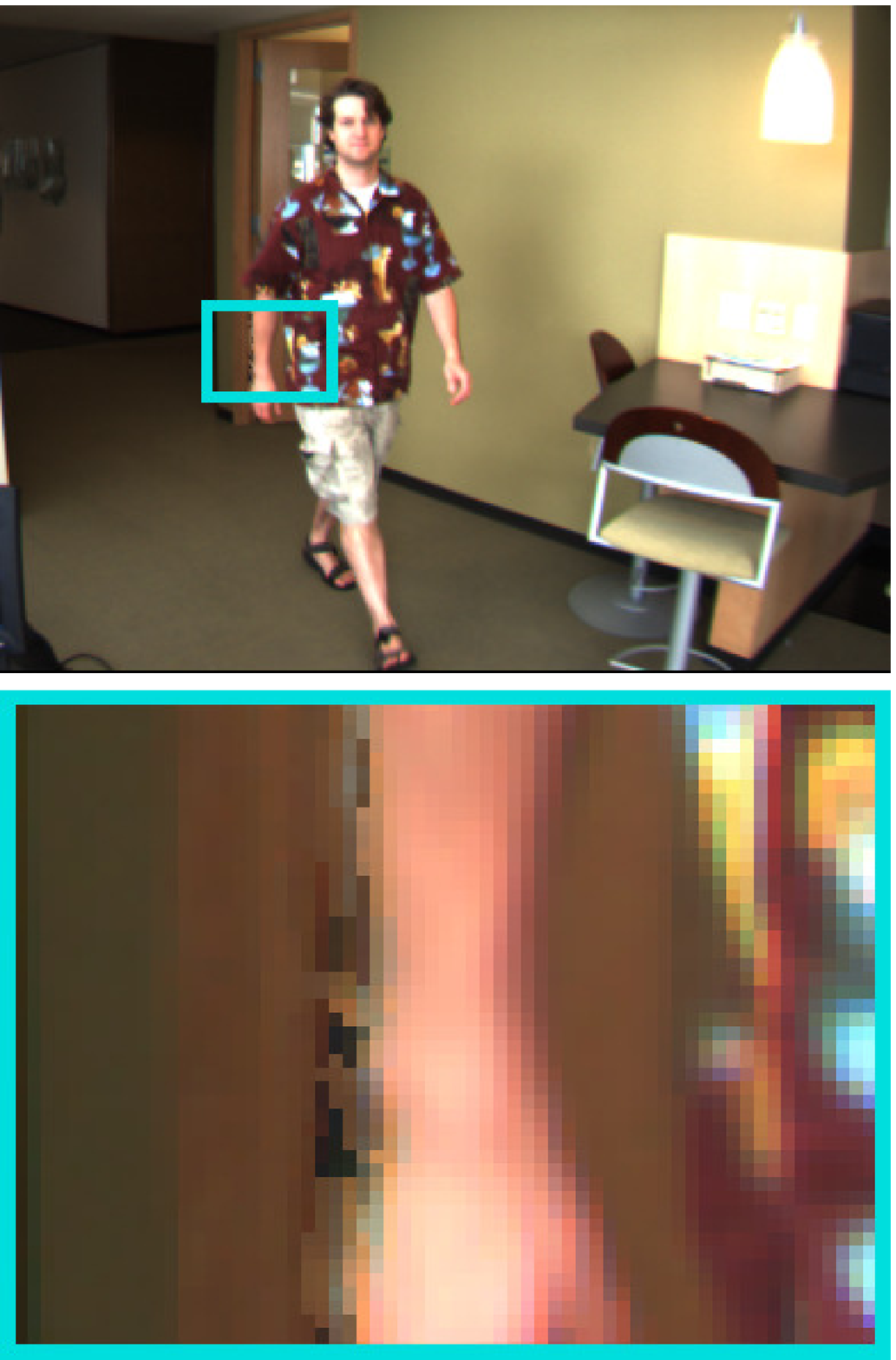}
        &
            \includegraphics[width=\itemwidth]{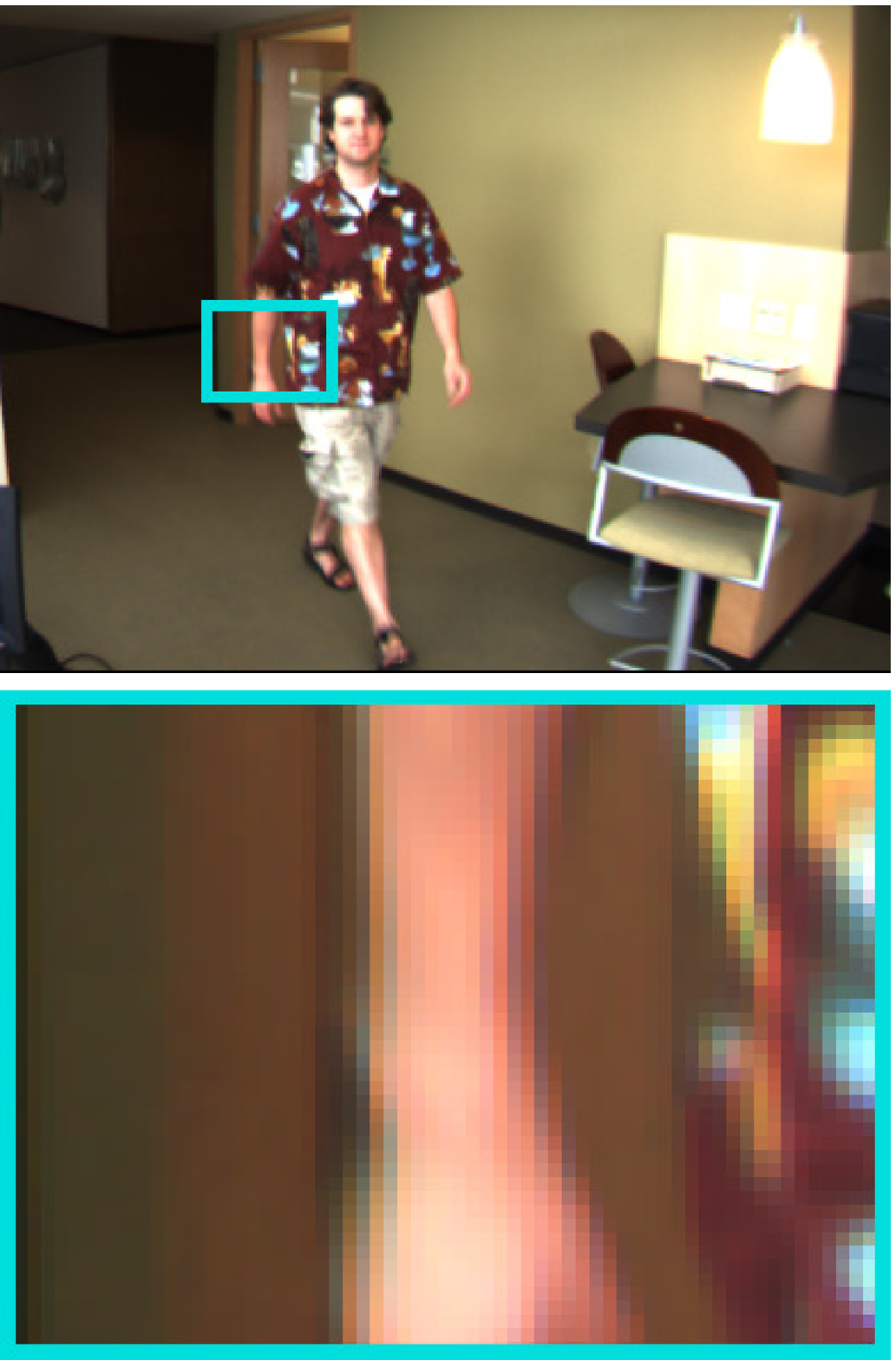}
        &
            \includegraphics[width=\itemwidth]{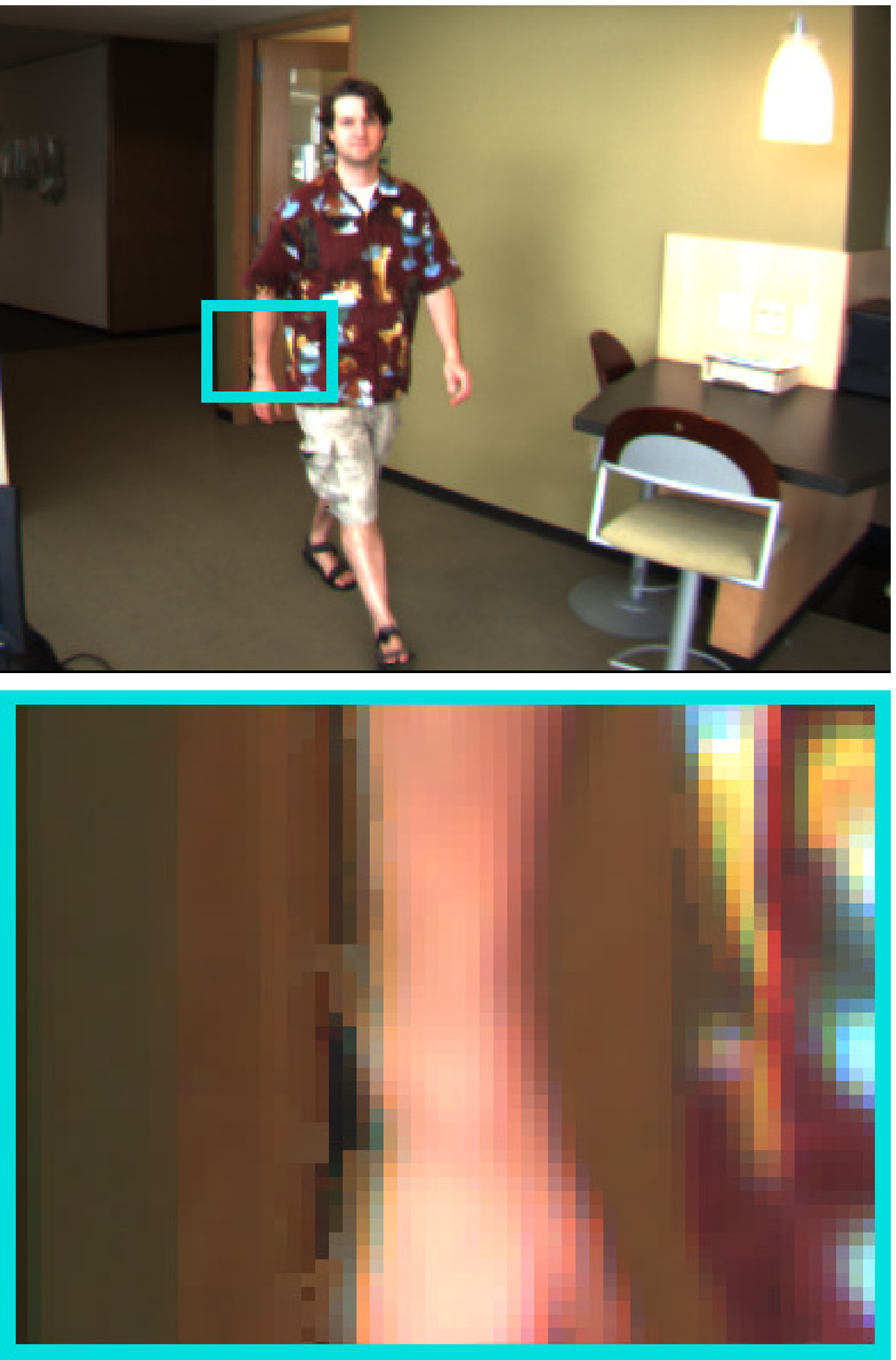}
        &
            \includegraphics[width=\itemwidth]{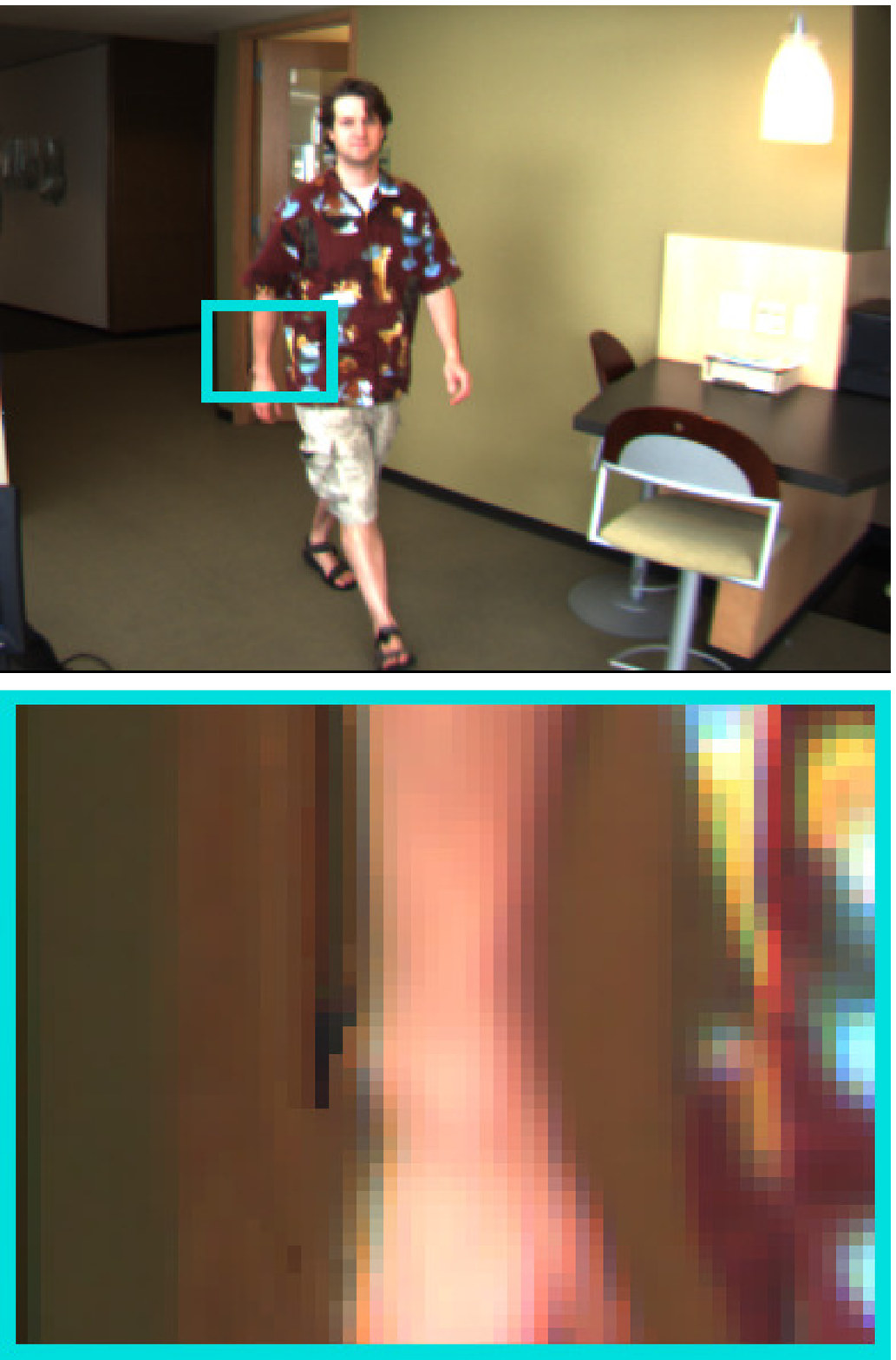}
        \vspace{-0.1cm} \\
            \footnotesize Input frame 1
        &
            \footnotesize Ours
        &
            \footnotesize Meyer~\etal
        &
            \footnotesize DeepFlow2
        &
            \footnotesize FlowNetS
        &
            \footnotesize MDP-Flow2
        &
            \footnotesize Brox~\etal
        \\
    \end{tabularx}\vspace{-0.1in}
    \caption{Qualitative evaluation with respect to occlusion.}\vspace{-0.2in}
    \label{fig:occlusion}
\end{figure*}

\noindent\textbf{Occlusion.} One of the biggest challenges for optical flow estimation is occlusion. When optical flow is not reliable or unavailable in occluded regions, frame interpolation methods need to fill in holes, such as by interpolating flow from neighboring pixels~\cite{Baker_OTHER_2011}. Our method adopts a learning approach to obtain proper convolution kernels that lead to visually appealing pixel synthesis results for occluded regions, as shown in Figure~\ref{fig:occlusion}.

\begin{figure}\centering
    \setlength{\tabcolsep}{0.0cm}

    \includegraphics[width=\columnwidth]{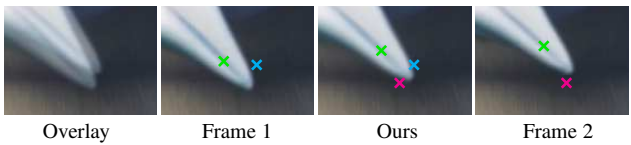}\vspace{-0.1cm}
    \begin{tabularx}{\columnwidth}{Y @{\hspace{0.2cm}} Y @{\hspace{0.2cm}} Y @{\hspace{0.2cm}} Y}
            \footnotesize Overlay
        &
            \footnotesize Frame 1
        &
            \footnotesize Ours
        &
            \footnotesize Frame 2
        \\
    \end{tabularx}\vspace{-0.1in}
    \caption{Occlusion handling.}\vspace{-0.2in}
    \label{fig:handling}
\end{figure}

To better understand how our method handles occlusion, we examine the convolution kernels of pixels in the occluded regions. As shown in Figure~\ref{fig:pipeline}, a convolution kernel can be divided into two sub-kernels, each of which is used to convolve with one of the two input patches. For the ease of illustration, we compute the centroid of each sub-kernel and mark it using \textbf{x} in the corresponding input patch to indicate where the output pixel gets its color. Figure~\ref{fig:handling} shows an example where the white leaf moves up from Frame 1 to Frame 2. The occlusion can be seen in the left image that overlays two input frames. For this example, the pixel indicated by the green \textbf{x} is visible in both frames and our kernel shows that the color of this pixel is interpolated from both frames. In contrast, the pixel indicated by the red \textbf{x} is visible only in Frame 2. We find that the sum of all the coefficients in the sub-kernel for Frame 1 is almost zero, which indicates Frame 1 does not contribute to this pixel and this pixel gets its color only from Frame 2. Similarly, the pixel indicated by the cyan \textbf{x} is only visible in Frame 1. Our kernel correctly accounts for this occlusion and gets its color from Frame 1 only.

\begin{figure}\centering
    \includegraphics[width=\columnwidth]{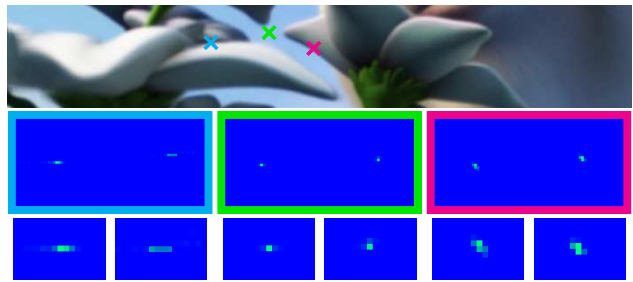}\vspace{-0.1in}
    \caption{Convolution kernels. The third row provides magnified views into the non-zero regions in the kernels in the second row. While our neural network does not explicitly model the frame interpolation procedure, it is able to estimate convolution kernels that enable similar pixel interpolation to the flow-based interpolation methods. More importantly, our kernels are spatially adaptive and edge-aware, such as those for the pixels indicated by the red and cyan \textbf{x}.}\vspace{-0.25in}
    \label{fig:kernel}
\end{figure}

\subsection{Edge-aware pixel interpolation}

In the above, we discussed how our estimated convolution kernels appropriately handle occlusion for frame interpolation. We now examine how these kernels adapt to image features. In Figure~\ref{fig:kernel}, we sample three pixels in the interpolated image. We show their kernels at the bottom. The correspondence between a pixel and its convolution kernel is established by color. First, for all these kernels, only a very small number of kernel elements have non-zero values. (The use of the spatial softmax layer in our neural network already guarantees that the kernel element values are non-negative and sum up to one.) Furthermore, all these non-zero elements are spatially grouped together. This corresponds well with a typical flow-based interpolation method that finds corresponding pixels or their neighborhood in two frames and then interpolate. Second, for a pixel in a flat region such as the one indicated by the green \textbf{x}, its kernel only has two elements with significant values. Each of these two kernel elements corresponds to the relevant pixel in the corresponding input frame. This is also consistent with the flow-based interpolation methods although our neural network does not explicitly model the frame interpolation procedure. Third, more interestingly, for pixels along image edges, such as the ones indicated by the red and cyan \textbf{x}, the kernels are anisotropic and their orientations align well with the edge directions. This shows that our neural network learns to estimate convolution kernels that enable edge-aware pixel interpolation, which is critical to produce sharp interpolation results.

\begin{figure}\centering
    \setlength{\tabcolsep}{0.0cm}
    \setlength{\itemwidth}{4.1cm}

    \begin{tabularx}{\textwidth}{c @{\hspace{0.05cm}} c}
            \includegraphics[width=\itemwidth]{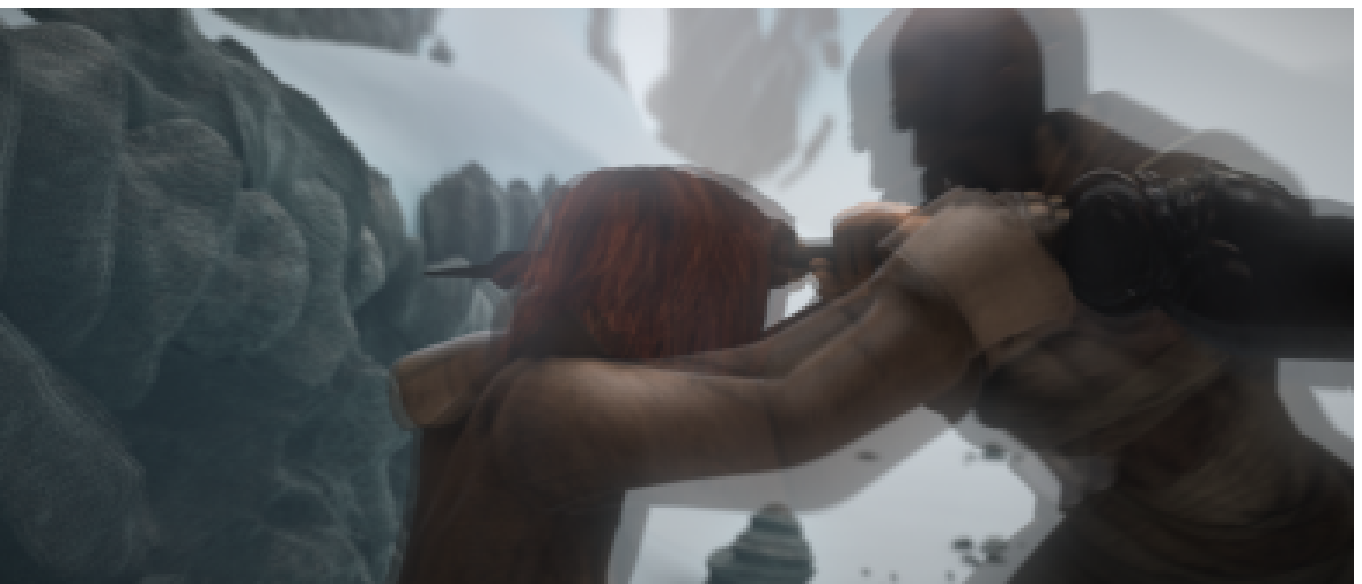}
        &
            \includegraphics[width=\itemwidth, height=1.75cm]{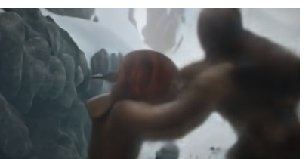}
        \vspace{-0.1cm} \\
            \footnotesize Overlayed input
        &
            \footnotesize Long~\etal
        \\
    \end{tabularx}
    \begin{tabularx}{\textwidth}{c @{\hspace{0.05cm}} c}
            \includegraphics[width=\itemwidth]{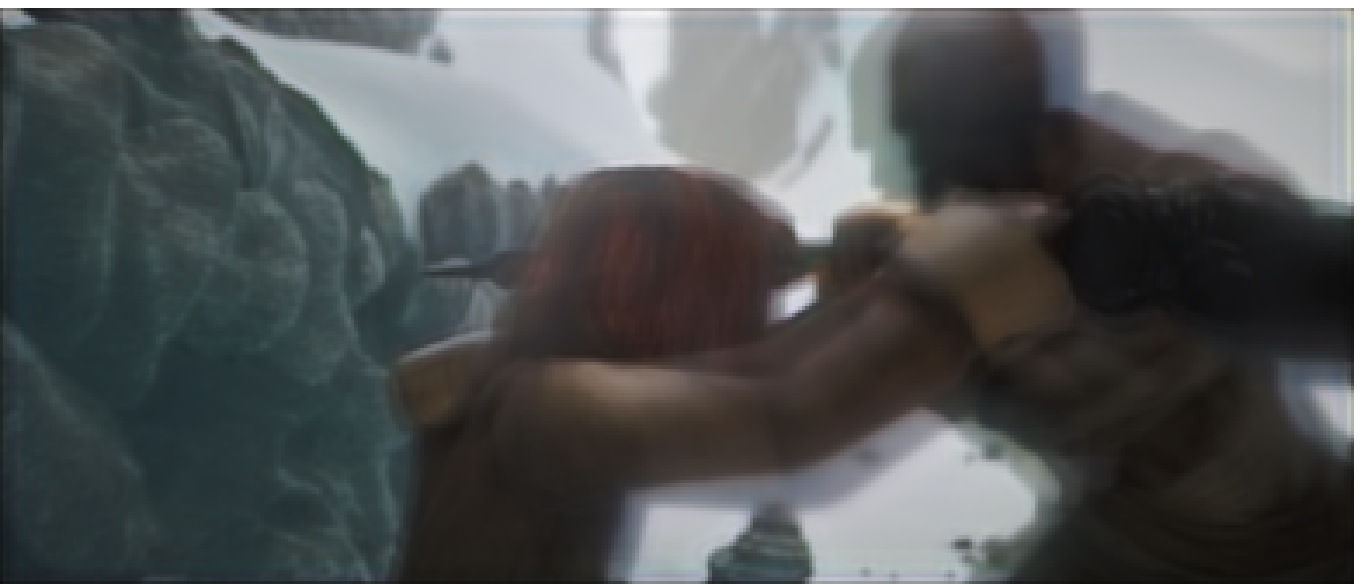}
        &
            \includegraphics[width=\itemwidth]{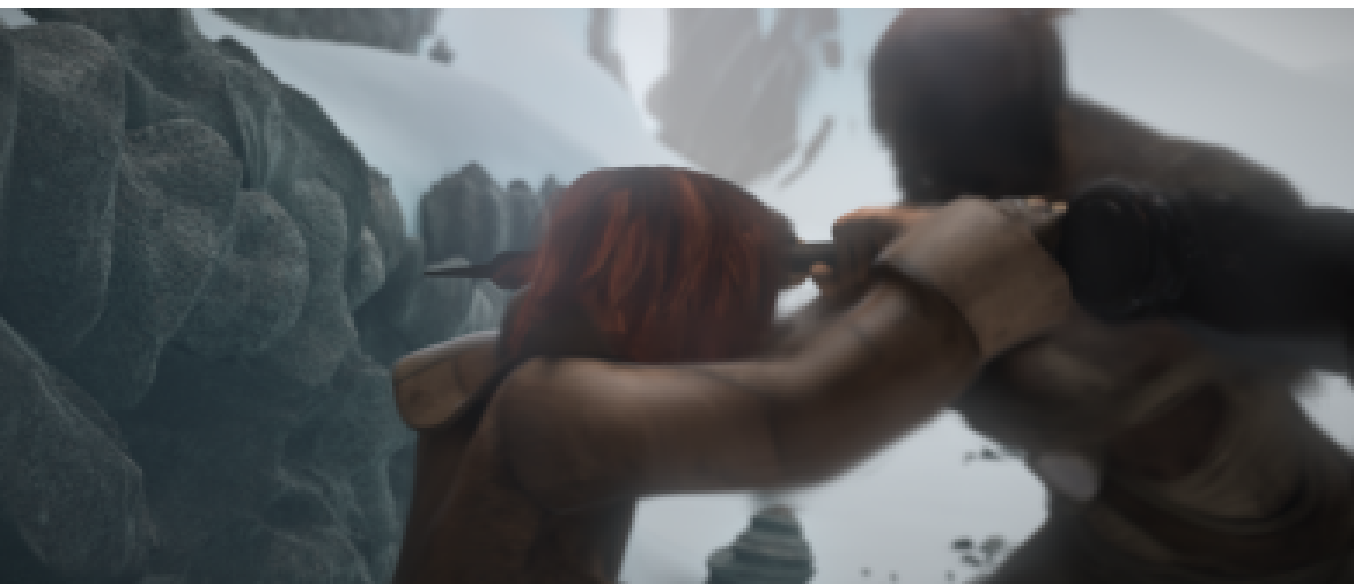}
        \vspace{-0.1cm} \\
            \footnotesize Direct
        &
            \footnotesize Ours
        \\
    \end{tabularx}\vspace{-0.3cm}
    \caption{Comparison with direct synthesis.}\vspace{-0.5cm}
    \label{fig:sintel}
\end{figure}

\subsection{Discussion} 

Our method is scalable to large images due to its pixel-wise nature. Furthermore, the shift-and-stitch implementation of our neural network allows us to both parallel processing multiple pixels and reduce the redundancy in computing the convolution kernels for these pixels. On a single Nvidia Titan X, this implementation takes about $2.8$ seconds with $3.5$ gigabytes of memory for a $640 \times 480$ image, and $9.1$ seconds with $4.7$ gigabytes for $1280 \times 720$, and $21.6$ seconds with $6.8$ gigabytes for $1920 \times 1080$.

We experimented with a baseline neural network by modifying our network to directly synthesize pixels. We found that this baseline produces a blurry result for an example from the Sintel benchmark~\cite{Butler_ECCV_2012}, as shown in Figure~\ref{fig:sintel}. In the same figure, we furthermore show a comparison with the method from Long~\etal~\cite{Long_ECCV_2016} that performs video frame interpolation as an intermediate step for optical flow estimation. While their result is better than our baseline, it is still not as sharp as ours.

\begin{figure}\centering
    \includegraphics[]{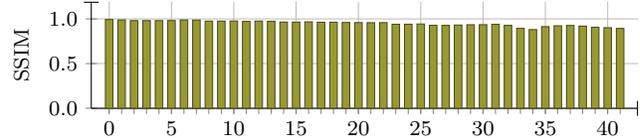}\vspace{-0.1in}
	\caption{Interpolation quality of our method with respect to the flow magnitude (pixels).}\vspace{-0.1in}
	\label{fig:magnitude}
\end{figure}

\begin{figure}\centering
    \setlength{\tabcolsep}{0.0cm}
    \setlength{\itemwidth}{4.1cm}

    \begin{tabularx}{\textwidth}{c @{\hspace{0.05cm}} c}
            \includegraphics[width=\itemwidth]{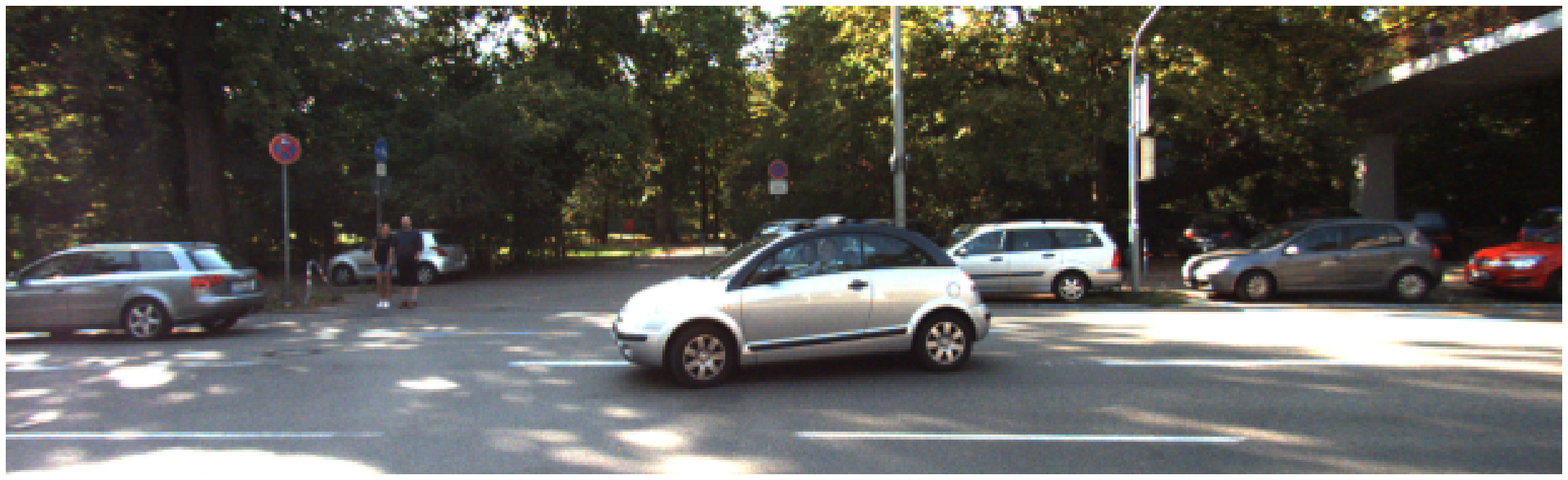}
        &
            \includegraphics[width=\itemwidth]{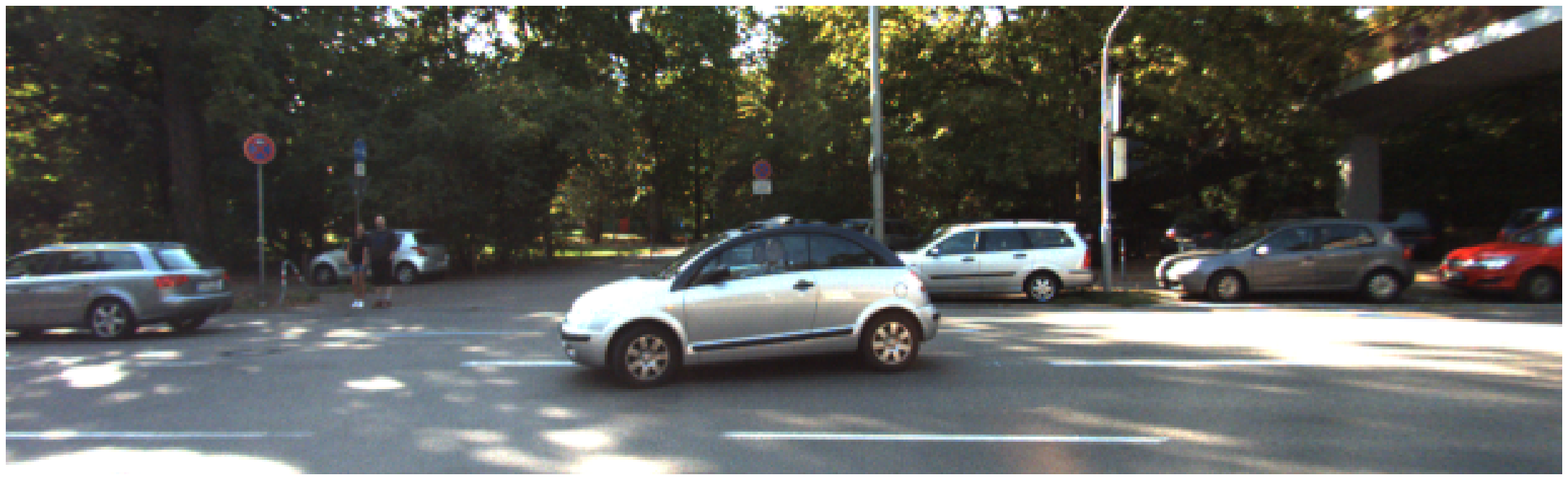}
        \vspace{-0.1cm} \\
            \footnotesize (a) Left view
        &
            \footnotesize (b) Right view
        \\
    \end{tabularx}
    \begin{tabularx}{\textwidth}{c @{\hspace{0.05cm}} c}
            \includegraphics[width=\itemwidth]{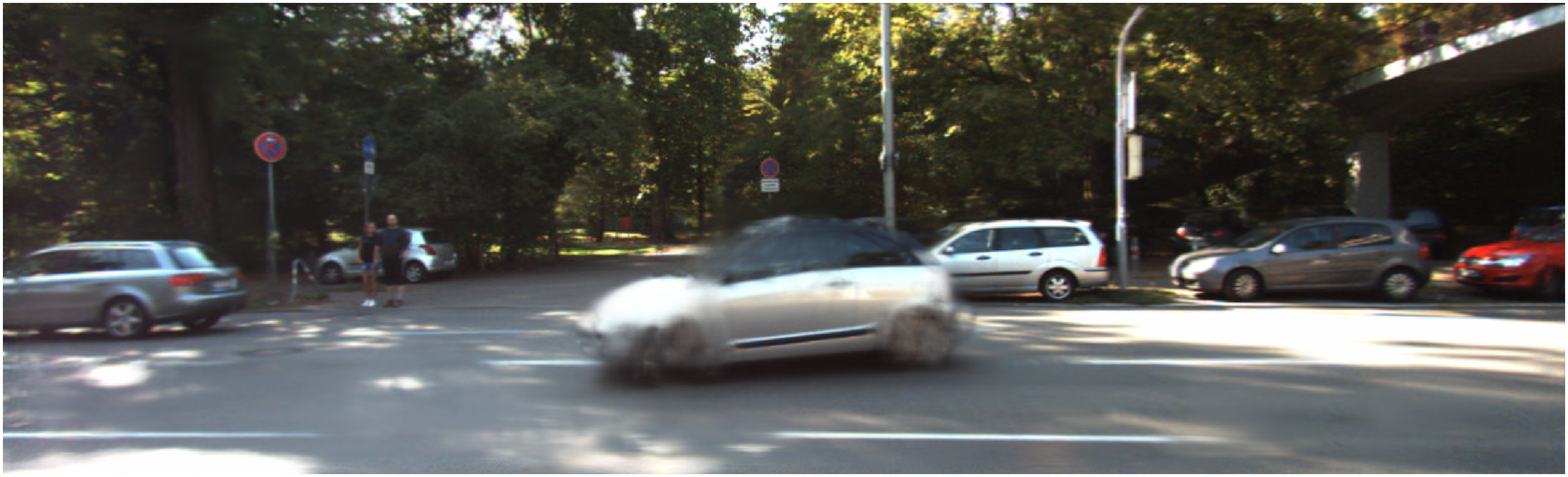}
        &
            \includegraphics[width=\itemwidth]{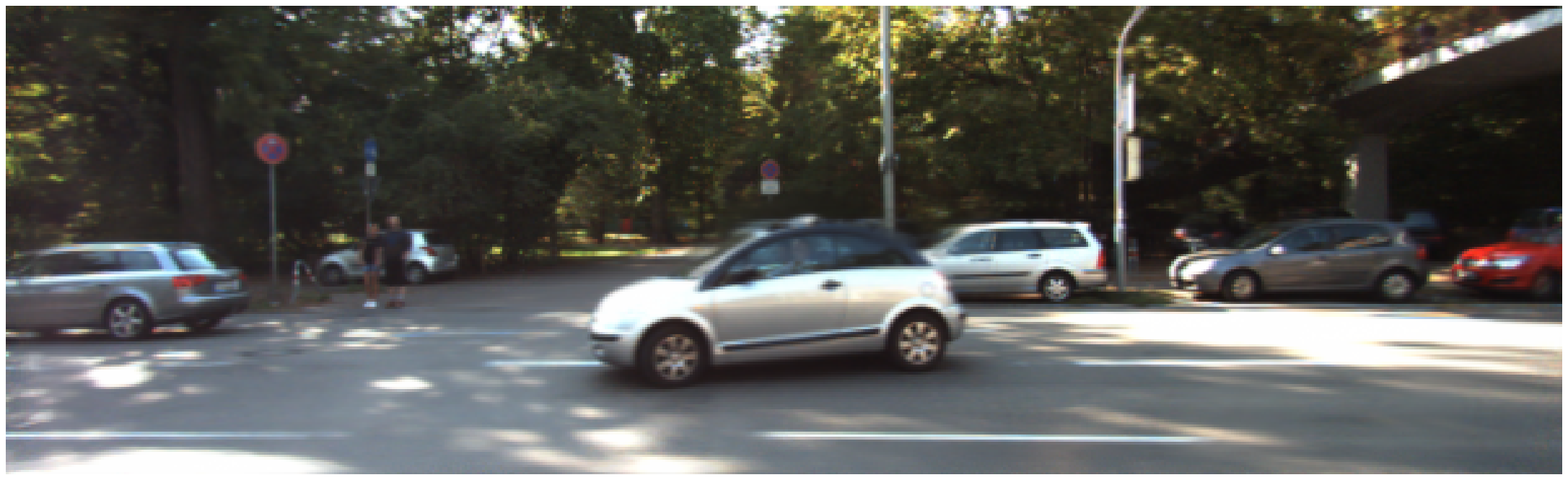}
        \vspace{-0.1cm} \\
            \footnotesize (c) Ours - full resolution
        &
            \footnotesize (d) Ours - half resolution
        \\
    \end{tabularx}\vspace{-0.3cm}
    \caption{Interpolation of a stereo image.}\vspace{-0.5cm}
    \label{fig:stereo}
\end{figure}

The amount of motion that our method can handle is necessarily limited by the convolution kernel size in our neural network, which is currently $41 \times 82$. As shown in Figure~\ref{fig:magnitude}, our method can handle motion within 41 pixels well. However, any large motion beyond 41 pixels, cannot currently be handled by our system. Figure~\ref{fig:stereo} shows a pair of stereo image from the KITTI benchmark~\cite{Menze_CVPR_2015}. When using our method to interpolate a middle frame between the left and right view, the car is blurred due to the large disparity (over 41 pixels), as shown in (c). After downscaling the input images to half of their original size, our method interpolates well, as shown in (d). In the future, we plan to address this issue by exploring multi-scale strategies, such as those used for optical flow estimation~\cite{Ranjan_CORR_2016}.

Unlike optical flow- or phased-based methods, our method is currently only able to interpolate a single frame between two given frames as our neural network is trained to interpolate the middle frame. While we can continue the synthesis recursively to also interpolate frames at $t = 0.25$ and $t = 0.75$ for example, our method is unable to interpolate a frame at an arbitrary time. It will be interesting to borrow from recent work for view synthesis~\cite{Dosovitskiy_CVPR_2015, Kalantari_TOG_2016, Kulkarni_NIPS_2015, Tatarchenko_ECCV_2016, Zhou_ECCV_2016} and extend our neural network such that it can take a variable as input to control the temporal step of the interpolation in order to interpolate an arbitrary number of frames like flow- or phase-based methods.

\section{Conclusion}
\label{sec:concl}

This paper presents a video frame interpolation method that combines the two steps of a frame interpolation algorithm, motion estimation and pixel interpolation, into a single step of local convolution with two input frames. The convolution kernel captures both the motion information and re-sampling coefficients for proper pixel interpolation. We develop a deep fully convolutional neural network that is able to estimate spatially-adaptive convolution kernels that allow for edge-aware pixel synthesis to produce sharp interpolation results. This neural network can be trained directly from widely available video data. Our experiments show that our method enables high-quality frame interpolation and handles challenging cases like occlusion, blur, and abrupt brightness change well.

\noindent\textbf{Acknowledgments.} The top image in Figure~\ref{fig:blurriness} is used with permission from Rafael McStan while the other images in Figures~\ref{fig:blurriness},~\ref{fig:brightness},~\ref{fig:occlusion} are used under a Creative Commons license from the Blender Foundation and the city of Nuremberg. We thank Nvidia for their GPU donation. This work was supported by NSF IIS-1321119.

{\small
\bibliographystyle{ieee}
\bibliography{egbib}
}

\end{document}